\pdfoutput=1
\documentclass[10pt,twocolumn,letterpaper]{article}
\usepackage[pagenumbers]{cvpr}

\usepackage{cuted} %

\usepackage{currfile} %

\usepackage{caption} %

\usepackage{multirow}
\usepackage{colortbl}
\definecolor{bestcell}{HTML}{DCE9F6}
\newcommand{\best}[1]{\cellcolor{bestcell}\textbf{#1}}
\definecolor{secondcell}{HTML}{EEF4FB}
\newcommand{\secondbest}[1]{\cellcolor{secondcell}\underline{#1}}
\newcommand{\ocve}{\textup{\textsc{OCVE}}}

\definecolor{cvprblue}{rgb}{0.21,0.49,0.74}
\usepackage[pagebackref,breaklinks,colorlinks,allcolors=cvprblue]{hyperref}
\def\confName{CVPR}
\def\confYear{2027}
\title{What You See Is What You Certify:\\Concept-Based Certificates for Unseen Compositions}
\author{Phuc Linh Ngo\textsuperscript{1} \quad Thang Nguyen Duc\textsuperscript{1} \quad Truong Thanh Hung Nguyen\textsuperscript{2} \quad Hung Cao\textsuperscript{2}\\[0.5ex]
\textsuperscript{1}Technische Universit\"at Darmstadt \qquad \textsuperscript{2}University of New Brunswick}
\begin{document}
\maketitle
\begin{strip}
\centering
\includegraphics[width=\textwidth]{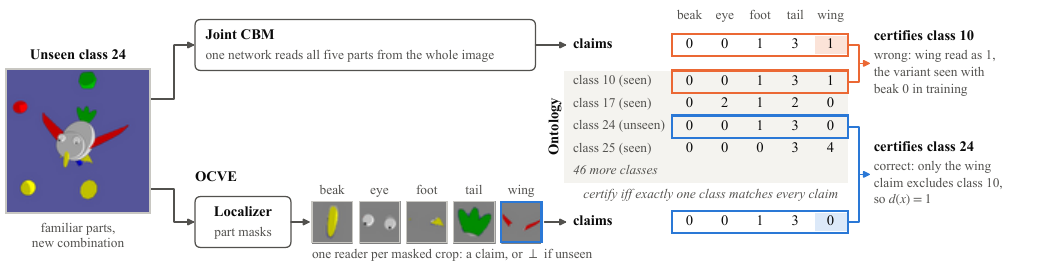}
\captionsetup{hypcap=false}
\captionof{figure}{\textbf{Seen, not inferred.} The ontology gives each class one variant per part; the unseen FunnyBirds class 24 differs from the seen class 10 only in its wing. A joint concept bottleneck model reads the wing from the whole image as the variant that co-occurs with the beak in training and certifies class 10. \ocve{} reads each part from its own masked crop and certifies class 24, the only class matching its claims; only the wing claim excludes class 10, so $d(x)=1$. The image is the first unseen-class test image, by index, on which the joint model certifies a wrong class through one copied part, and \ocve{} certifies the true one (49 of 2{,}100 qualify).}
\captionsetup{hypcap=true}
\label{fig:\currfilebase}%
\end{strip}

\begin{abstract}
Concept-based models justify a decision by concepts they claim to see, decoded through an explicit ontology. This justification assumes that each predicted concept reflects what is visible in its region, and a growing literature shows that the assumption fails: concept predictions leak label information, do not respect locality, and are poorly grounded on new attribute combinations. A decision decoded from such concepts can satisfy the ontology and still be wrong, so we ask when it can serve as a certificate. On classes that combine familiar parts in unfamiliar ways, a joint concept bottleneck model and a concept embedding model, trained as is standard, decode with the correct ontology yet certify wrong, mostly familiar, classes 93--94\% of the time (60\% for a joint model trained only on visible parts and gated by a part localizer). The joint model's part errors copy co-occurring values more often than chance, which we call cross-concept shortcutting. Treating the ontology as an error-detecting code, we show that a certificate needs visible evidence that excludes every rival, and unseen parts reported as such, and that misreads must be rare and not driven by a shared shortcut. A minimal reference design, \ocve{}, reads each part from a local crop in its own forward pass. On unseen FunnyBirds classes, whose ontology is correct by construction, it certifies 52\% of images at 0.4\% risk, against 14\% and 33\% risk for mask-pooled features and full-image part readers (5.6\% when trained and gated in the same way) at similar zero-shot accuracy. The best baselines, even calibrated or restricted to unseen classes, and a decoder that only counts agreeing parts remain about 3--8 times riskier. On CUB, isolated region readers are more accurate than an attribute-prototype model (63.1\% vs.\ 55.3\%) and certify unseen species with less risk than features pooled in the same keypoint boxes (about 10 points) and than the prototype model (about 6 points). On real traffic signs, isolated readers lower certified risk from 99.8\% to 67\%, which is still far too high; most remaining errors are misread pictograms. Certificate reliability rests on the part readers; on seen classes the ontology brings no gain over a fairly trained black box, and isolation costs accuracy.
\end{abstract}

\section{Introduction}
\label{sec:intro}
Concept bottleneck models (CBMs)~\cite{koh2020cbm} and attribute-based recognizers~\cite{lampert2009dap} make two promises. Their explanations name the evidence behind a decision (``beak type 2, tail type 7''), and an explicit concept-to-class map lets them recognize classes described only by a specification. Both promises assume that each predicted concept reflects what is visible in its region. A growing literature shows the assumption fails: concept predictions leak label information~\cite{havasi2022leakage,parisini2025leakage}, do not respect locality~\cite{raman2025locality}, and are poorly grounded on new attribute combinations~\cite{bader2025sub}.

We study \emph{when an ontology can turn a model's concept predictions into a certificate one can trust}. We derive what a certificate requires, build a minimal model that meets these requirements, remove its components one at a time, and find where its certificates stop being reliable.

\textbf{(i) The ontology must carry information the data does not.} On classes seen in training, a fairly trained black box absorbs the concept-to-class map; we find no accuracy or robustness gain from the ontology there (Sec.~\ref{sec:seen}). On \emph{unseen compositions}, the ontology alone defines the class.

\textbf{(ii) Evidence should be isolated.} A reader that sees the whole image can infer a part from its context (Fig.~\ref{fig:teaser}). Under a compositional shift, these inferences are systematically wrong yet mutually consistent, so the joint CBM reproduces the code of a familiar class from inferred part values and certifies it. We call this \emph{cross-concept shortcutting} and measure it directly (Sec.~\ref{sec:mechanism}). Full-image part readers trained on visible parts show a small average shortcut ($\lambda\le0.008$) yet are about ten times riskier than local readers; for wrong readings, their attribution maps mostly point elsewhere, although this evidence is only correlational (Sec.~\ref{sec:ablate}).

\textbf{(iii) Reliability depends on the evidence in each image.} On FunnyBirds, all 22 wrong \ocve{} certificates had $d(x)=1$, and none of the 1{,}195 at $d(x)\ge2$ was wrong, although these images also claim more parts (Sec.~\ref{sec:stress}). The count helps only when the code leaves a margin and claims fail independently: on traffic signs, nearly every unseen code has a neighbor at distance 1, and all 29 \ocve{} certificates of unseen signs with $d(x)=2$ were wrong; in 16 of them, one reader misread both pictogram and text (appendix).

\paragraph{Contributions.}
\begin{itemize}
\item \textbf{Audit.} A joint CBM and a concept embedding model decoding with the correct ontology certify wrong classes 93--94\% of the time on unseen compositions. One measurable cause is cross-concept shortcutting, measured by $\lambda$ (0.055 for the joint CBM, $-0.004$ for \ocve{}); erasing or swapping parts, or visible-part training with localizer gating, leaves the joint CBM's risk at 60--93\% (Sec.~\ref{sec:mechanism}--\ref{sec:ablate}).
\item \textbf{Conditions.} We give a coding-theoretic account of ontology-based certification with abstention. It states what a certificate needs (visible claims excluding every rival, and unseen parts reported as $\bot$), when its error bound is informative (rare misreads not driven by a shared shortcut), and how much evidence supports it (a per-image evidence distance). It also implies that minimal sufficient explanations tolerate no misread concept (Sec.~\ref{sec:framework}).
\item \textbf{Reference design.} We build \ocve{}, a minimal design meeting these conditions: a class-agnostic localizer, one reader per part that sees only its own crop and reports $\bot$ when its part is not found or is read with low confidence, and an ontology decoder that certifies or abstains. In ablations, the essential ingredient is a local window read in its own forward pass; per-part networks, masking, centering, and magnification are not required, and neither mask pooling nor augmentation reproduces it (Sec.~\ref{sec:results}).
\item \textbf{Boundaries.} On FunnyBirds, certificates need a decoder that excludes rivals and a correct ontology. On seen classes, an apparent robustness gain came entirely from oracle masks. On traffic signs, isolated readers lower certified risk only to 67\%. On CUB, isolated readers lead a prototype model in risk by about 6 points but lose this lead when the head is hidden (Sec.~\ref{sec:mechanism}--\ref{sec:cub}).
\end{itemize}

\section{Related work}
\paragraph{Concept bottlenecks and leakage.} CBMs~\cite{koh2020cbm} predict human concepts and then the label. Soft concept scores leak task information~\cite{mahinpei2021promises,havasi2022leakage}; leakage has been quantified information-theoretically~\cite{parisini2025leakage}, and a controlled form of it can support accuracy and intervention when the concept set is incomplete~\cite{espinosa2026defense}. We study a related failure, in which a concept value is inferred from other parts and used as a premise of a certificate. Strongly correlated concepts are hard to tell apart~\cite{heidemann2023correlation}, and in neuro-symbolic models, concepts with unintended meanings can satisfy the prior knowledge and still yield correct labels (reasoning shortcuts~\cite{marconato2023reasoning}). Our concepts are supervised, yet on unseen compositions, misread concepts still satisfy the ontology. Many CBMs ignore concept locality~\cite{raman2025locality}, and concurrent work on CUB finds that local patches make grounding testable without guaranteeing it~\cite{saralajew2026auditing}. Prototype-grounded concept models expose what each concept means through learned visual prototypes~\cite{colamonaco2026pgcm}; we ask whether certificates built on such concepts survive unseen compositions. Concept embedding models~\cite{espinosa2022cem} replace scalar concept scores with embeddings to recover task accuracy; on unseen compositions they show cross-concept shortcuts like a joint CBM. Spatial variants ground concepts through attention or part queries over a shared backbone~\cite{panousis2024coarse,benou2025salf,ramachandram2026pfcbm}; we test the simplest such grounding, mask pooling of late-layer features, and its certificates are still wrong 13.8\% of the time on unseen compositions. SUB~\cite{bader2025sub} evaluates CBMs on substituted attributes, and we test our readers on it.
\paragraph{Attribute-based and compositional zero-shot learning.} DAP~\cite{lampert2009dap} and semantic output codes~\cite{palatucci2009zero} decode classes from attribute predictions, and error-correcting output codes~\cite{dietterich1995ecoc} decode them from binary classifiers through a designed code. Here the ontology fixes the code, which we use to detect misreads and abstain rather than to correct them. Attribute decorrelation by semantic groups~\cite{jayaraman2014decorrelating} reduces features shared between attributes, region-based zero-shot models localize attributes with attention~\cite{xu2020apn,huynh2020dazle,chen2022transzero} or detected parts~\cite{elhoseiny2017link}, and fine-grained recognizers classify cropped or pose-normalized parts~\cite{zhang2014partrcnn,branson2014pose,huang2016partstacked}; we ask when part crops make a certificate reliable rather than whether they raise accuracy. Compositional zero-shot learning~\cite{mancini2021open,nayak2023csp} targets unseen attribute--object pairs in photos.
\paragraph{Formal and verifiable explanations.} Abductive and contrastive explanations are dual through minimal hitting sets~\cite{ignatiev2020contrastive}; robust and probabilistic variants perturb features outside the explanation~\cite{lamalfa2021robust,wang2021probabilistic,izza2024distance}. We study misreads \emph{inside} the explanation; the combinatorics of tolerating them are known from identifying codes and test design~\cite{karpovsky1998identifying,honkala2007robust}.
\paragraph{Selective prediction and evaluation of XAI.} Explanations often raise trust without helping users detect errors~\cite{kim2022hive,nguyen2021effectiveness}. Probabilistic CBMs attach an uncertainty to each concept~\cite{kim2023probcbm}, and conceptual safeguards abstain when propagated uncertainty is high~\cite{joren2024safeguards}. Our certificates abstain unless the claims single out one class, and we ask when the claims deserve trust. We compare certificates with confidence ranking at matched coverage, the standard selective-prediction baseline~\cite{geifman2017selective}, also after calibrated stacking for generalized zero-shot learning~\cite{chao2016empirical}. FunnyBirds~\cite{hesse2023funnybirds} provides part-level interventions that we use for stress tests.

\section{Ontologies as error-detecting codes}\label{sec:framework}
\paragraph{Setup.} An ontology assigns each class $y$ a codeword $C(y)=(C_g(y))_{g\in\mathcal G}$ over evidence groups $\mathcal G$ (parts), where $C_g(y)\neq\emptyset$ is the set of values class $y$ allows for group $g$. A perception front-end returns for each group a claim $\hat v_g$ or $\bot$ (unseen). With $O=\{g:\hat v_g\neq\bot\}$, the decoder computes the compatible set $D(O)=\{y:\hat v_g\in C_g(y)\ \forall g\in O\}$ and \emph{certifies} $\hat y$ iff $D(O)=\{\hat y\}$, otherwise abstains or returns $D(O)$. A certificate is a logical statement with no statistical guarantee; its empirical error rate is the certified risk.
\paragraph{Evidence distance.} For a rival $y'$, let $R(y')=\{g\in O:\hat v_g\notin C_g(y')\}$ be the observed claims that exclude it; the evidence distance is $d(x)=\min_{y'\neq\hat y}|R(y')|$.
\paragraph{Tolerance.} An explanation $E\subseteq O$ is $k$-tolerant if no rival becomes consistent after changing at most $k$ of its claims, \ie, $|E\cap R(y')|\ge k+1$ for all rivals, a $(k+1)$-multicover of the rival-exclusion sets~\cite{berge1989hypergraphs,karpovsky1998identifying} (for $k=0$, the abductive/contrastive hitting-set duality~\cite{ignatiev2020contrastive}). Hence \emph{every subset-minimal sufficient explanation has tolerance zero}: each claim is a single point of failure, so explanation criteria for perception should favor tolerance over minimality.
\paragraph{What the bound needs.} Assume the ontology is correct (Sec.~\ref{sec:ablate} tests wrong entries) and contains the true class $y^\ast$. A wrong certificate excludes $y^\ast$, so every claim in $R(y^\ast)$ is a misread, and because $|R(y^\ast)|\ge d(x)$, a wrong certificate implies at least $d(x)$ misreads. This needs no independence assumption. A numerical risk bound also needs a model of how misreads co-occur and of which groups get claimed; for fixed $d$ and $m$ fixed claims that err independently with probability at most $\varepsilon$, $P(\text{wrong certificate},\,d(x)\ge d)\le\binom{m}{d}\varepsilon^{d}$, a joint probability rather than the risk among certified images. The count is informative only when misreads are rare and not driven by a common cause, and only for classes that differ from every rival in at least $d$ claimed groups, since for other classes every correct certificate has $d(x)<d$. Cross-concept shortcuts are such a cause: a reader that infers part $j$ from part $i$ turns several misreads into one event. Reading parts from disjoint pixels is one way to avoid them, but in our ablations it is not required (Sec.~\ref{sec:ablate}). We quantify shortcuts by
\begin{equation}
\lambda_{j\leftarrow i}=P\big(\hat a_j=\tilde a_j(a_i)\big)-P\big(\hat a_j=\tilde a_j(a_i)\,\big|\,i\text{ occluded}\big),
\end{equation}
where $\tilde a_j(a_i)$ is the value of $j$ co-occurring most with $a_i$ in training classes, on test images that show both parts and whose true $a_j$ differs from it; $\lambda$ averages all ordered pairs.

\section{\texorpdfstring{\ocve{}}{OCVE}: ontology-coded verifiable explanations}\label{sec:method}
\begin{figure}[t]\centering
\includegraphics[width=\linewidth]{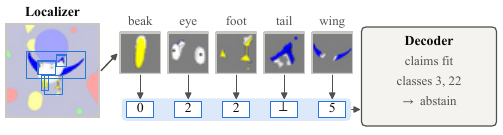}
\caption{\ocve{} on an unseen-class image without its tail (class 3; real outputs; selection rule: appendix). The tail reader reports $\bot$, the other claims fit classes 3 and 22, and the decoder abstains; the joint CBM reads a tail and certifies class 22.}
\label{fig:method}
\end{figure}
\paragraph{Localizer.} A class-agnostic part segmenter (ResNet-18 trunk with an FPN head) predicts one mask per part group (Fig.~\ref{fig:method}). It localizes parts but does not predict their values.
\paragraph{Isolated readers.} For each group, the reader receives the bounding-box crop of the predicted mask with every pixel outside the mask set to gray, in a separate forward pass (ResNet-18, $96\times96$). Readers are trained on crops from predicted masks and include an \textsc{absent} class for regions that are not truly that part. A part is reported as $\bot$ when the localizer finds no mask for it or the reader's confidence falls below a threshold $\tau$ set per group. Five $96^2$ crops amount to 92\% of the pixels of one $224^2$ image; with the localizer, inference costs at most $2.4\times$ the FLOPs of one ResNet-18 pass.
\paragraph{Decoder.} The ontology decoder returns a certificate when one class remains and the compatible set otherwise; a soft decoder ($\arg\max_y\sum_g\log p_g(C_g(y))$) gives a full-coverage prediction for accuracy metrics. A certificate reports $d(x)$ and the claims that exclude the nearest rival.

\section{Protocol}\label{sec:protocol}
\paragraph{Reporting.} All four CUB runs are reported (Sec.~\ref{sec:cub} and appendix). Analyses decided after seeing a result are post hoc; the appendix lists them all, including every chance level, interval, attribution map, and error breakdown.
\paragraph{Data.} FunnyBirds~\cite{hesse2023funnybirds} contains 50 classes, each a unique combination of five part types with 4, 3, 4, 9, and 6 variants, rendered at $256^2$ with part segmentation. For unseen compositions, we use 3 splits that each hold out 10 classes while every part variant remains present among the seen classes. Each split has 12{,}000 training, 2{,}100 unseen-class test, and 1{,}000 seen-class test images. FunnyBirds removes one to five parts from 41\% of its training images~\cite{hesse2023funnybirds}; most of our test images come from that partition, and 41\% of them lack at least one part (Sec.~\ref{sec:mechanism} and the appendix analyze complete birds). On CUB-200-2011~\cite{wah2011cub}, we use 3 random splits of 50 unseen species and the proposed split~\cite{xian2019zsl}, with region-local attributes and predicted keypoints; 3 further random splits serve only the matched-keypoint and seed comparisons. MTSD~\cite{ertler2020mtsd} has 244 sign codes over five groups; each of 3 splits holds out 49 (appendix).
\paragraph{Ontologies.} The FunnyBirds ontology is the generator's class definitions (one variant per part), correct by construction. CUB's class-level attribute percentages, averaged over annotators~\cite{wah2011cub}, are noisy and not discrete, hence the stability rule of Sec.~\ref{sec:cub}. The MTSD table was derived from the label names by rule and verified against montages of validation crops per label, which corrected 728 of 2{,}000 entries (12 labels without validation crops, with 24 training signs, remain unchecked; appendix).
\paragraph{Baselines.} All baselines use the same backbone, data, and number of epochs. B1 is a black-box ResNet-18, which cannot predict unseen classes. B2 is a joint CBM with five heads on the full image and the same decoder as \ocve{}. B2-ind trains one ResNet-18 per part, each on the full image. B2-maskpool pools the late-layer features ($14\times14$) of a shared trunk inside each predicted part mask (soft isolation, as in part-query CBMs). APN-lite is our re-implementation of the attribute prototype network~\cite{xu2020apn} on ResNet-18. CEM is a concept embedding model~\cite{espinosa2022cem} with one 16-dimensional embedding per part variant, a task head on seen classes, and random concept intervention ($p=0.25$), decoded like B2. B2, B2-ind, CEM, and APN-lite learn class-level concept values for every part, as is standard~\cite{koh2020cbm}, including parts the image does not show. The readers of \ocve{} and B2-maskpool learn only the parts their localizer finds (Sec.~\ref{sec:ablate} retrains the other models on visible parts only). All recognition models use ResNet-18~\cite{he2016resnet}; ResNet-50 and ViT-B/16~\cite{dosovitskiy2021vit} replace it in backbone checks (Sec.~\ref{sec:ablate}). The upper block of Table~\ref{tab:unseen} reports the main training run. Table~\ref{tab:ablate}, Fig.~\ref{fig:isolation}, and the stress tests use a second training run (\ocve{}: 0.5\% vs.\ 0.4\% risk).
\paragraph{Metrics.} We report certified risk and coverage, and accuracy on unseen-class images over all 50 classes (GZSL) and among unseen classes (ZSL). For comparison with selective prediction, we also report the risk of confidence ranking at matched coverage. Stress tests measure the unwarranted-commitment rate under decision-critical occlusion, risk by $d(x)$, and set-valued coverage at fixed risk (appendix); the mechanism analysis measures the shortcut rate and $\lambda$.

\section{Results}\label{sec:results}
\subsection{Unseen compositions}\label{sec:unseen}
\begin{figure}[t]\centering
\includegraphics[width=\linewidth]{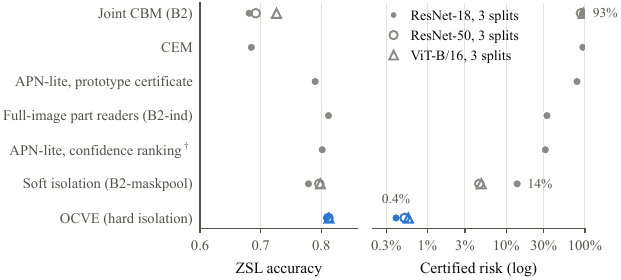}
\caption{Zero-shot accuracy hides certificate failure (FunnyBirds, unseen classes). Accuracy is similar across designs; certified risk spans over two orders of magnitude (means over 3 splits). $^\dagger$As in Table~\ref{tab:unseen}.}
\label{fig:divergence}
\end{figure}

\begin{table}[t]\centering\small
\caption{Unseen FunnyBirds classes (means over 3 random splits of 10 unseen classes). ZSL: accuracy among unseen classes. GZSL: accuracy over all 50 classes. Cert.\ risk: error rate among certified images, with coverage in parentheses (\ocve{}: 14 wrong of 3{,}274, 95\% interval 0.23--0.72\%). Seen classes: accuracy on all seen-class test images, reported for reference and not ranked; 42\% of them lack parts (perfect readers: 0.72). In the upper block, \ocve{} and B2-maskpool learn concept labels only for parts their localizer finds, and the others, as is standard, learn them for every part. The lower block (second run; Sec.~\ref{sec:ablate}) retrains these models on visible parts only, scores them with the localizer gate, and repeats \ocve{}. Shading per block: dark and bold, best per unseen-class column; light and underlined, second best. $^\dagger$Post hoc, second run: APN-lite's most confident images at \ocve{}'s coverage.}
\label{tab:unseen}
\setlength{\tabcolsep}{3pt}
\resizebox{\columnwidth}{!}{%
\begin{tabular}{lccc@{\hspace{10pt}}c}
\toprule
& \multicolumn{3}{c}{Unseen classes} & Seen classes\\
\cmidrule(lr){2-4}\cmidrule(l){5-5}
Model & GZSL$\uparrow$ & ZSL$\uparrow$ & Cert.\ risk$\downarrow$ (cov.) & Acc.\ (ref.)\\
\midrule
B2 joint CBM & 0.087 & 0.681 & 93.3\% (46\%) & 0.696\\
CEM & 0.101 & 0.685 & 94.2\% (46\%) & 0.697\\
B2-ind (full-image readers) & 0.512 & \best{0.812} & 33.0\% (32\%) & 0.730\\
B2-maskpool (soft isolation) & \secondbest{0.552} & 0.779 & \secondbest{13.8\% (37\%)} & 0.599\\
APN-lite (proto.\ certificate) & 0.499 & 0.790 & 79.4\% (29\%) & 0.697\\
APN-lite (confidence)$^\dagger$ & 0.480 & 0.802 & 31.5\% (52\%) & --\\
\ocve{} (hard isolation) & \best{0.678} & \secondbest{0.809} & \best{0.4\% (52\%)} & 0.605\\
\midrule
\multicolumn{5}{l}{\emph{Visible-part labels and localizer gate (post hoc)}}\\
B2 joint CBM & 0.302 & 0.788 & 60.1\% (36\%) & --\\
CEM & 0.246 & 0.765 & 71.7\% (37\%) & --\\
APN-lite (proto.\ certificate) & 0.471 & 0.805 & 39.7\% (34\%) & --\\
B2-ind (full-image readers) & \best{0.698} & \best{0.841} & \secondbest{5.6\% (49\%)} & --\\
\ocve{} (hard isolation) & \secondbest{0.677} & \secondbest{0.811} & \best{0.52\% (52\%)} & --\\
\bottomrule
\end{tabular}}
\end{table}
All models decode the same ontology, so each can in principle recognize an unseen class from its codeword. In the upper block of Table~\ref{tab:unseen}, zero-shot accuracy among unseen classes is 0.78--0.81 for every design except the joint CBM and CEM (0.68). Certified risk, however, differs widely. When \ocve{} certifies a class, it is wrong 0.4\% of the time, against 14\% for B2-maskpool, 33\% for B2-ind, whose separate readers still see the whole image, 79\% for APN-lite's prototype certificate, and 93\% for the joint CBM. It also certifies more images (52\% vs.\ 29--46\%). GZSL accuracy, which counts fallbacks to seen classes, broadly follows the same order. Certified risk falls monotonically with the strength of isolation in each split (\ocve{} 0.3--0.6\%, B2-maskpool 12--18\%, B2-ind 27--40\%) and with ResNet-50 and ViT-B/16 (Fig.~\ref{fig:divergence}). In the lower block, B2-ind falls below B2-maskpool's 13.8\% yet stays 11 times riskier than \ocve{}.

\paragraph{Confidence ranking (post hoc).} The strongest baselines we found are about 3--8 times riskier than \ocve{}, even with calibrated stacking or with decoding restricted to unseen classes. Ranking predictions by confidence is a stronger baseline than APN-lite's own certificate. At \ocve{}'s coverage, the most confident predictions over all 50 classes are wrong 17\% of the time for B2-maskpool, 24\% for B2-ind, 31.5\% for APN-lite, 92\% for CEM, and 93\% for the joint CBM. Concept models rank by the class probability propagated from their concepts, as in conceptual safeguards~\cite{joren2024safeguards}. With calibrated stacking~\cite{chao2016empirical} tuned on the test images (an optimistic choice), APN-lite reaches 3.2--3.7\% on pooled seen and unseen images (\ocve{}, uncalibrated: 0.47\%). With decoding restricted to unseen classes, as on CUB, \ocve{} reaches 0.36\% at 71\% coverage, B2-maskpool's certificates reach 1.9\% at 55\%, B2-ind's reach 1.0--1.2\% at 50--57\% with the localizer gate, and APN-lite reaches 3.0\%, whereas the joint CBM certifies 4.8\% of images at 26\% risk. Over all 50 classes, 99.8\% of a retrained joint CBM's wrong certificates name a seen class; the decoder is not told which classes are unseen (appendix).

\subsection{Mechanism: cross-concept shortcuts}\label{sec:mechanism}
On seen classes the joint CBM reads parts almost perfectly (0.94--0.99 per part); on unseen classes its accuracy drops to 0.26 (tail), 0.72 (beak), 0.77 (eye), and 0.81 (wing), while \ocve{}'s readers stay at 0.98--0.996. On complete birds, from which FunnyBirds removed no parts (second training run), the joint CBM reaches 0.93 seen-class accuracy (perfect readers: 0.96), yet 90\% of its unseen-class certificates are wrong. Training for 30 epochs (2.8 times \ocve{}'s training FLOPs) raises this accuracy to 0.95 (joint CBM) and 0.94 (CEM), whose risks without the gate stay at 94\% and 91\% (post hoc; appendix). Of the joint CBM's part errors on unseen classes in the second training run, 75\% copy the value that co-occurs most with another part of the class among seen classes (60\% expected from variant frequencies; \ocve{}: 64\% vs.\ 62\%). Because this rate depends on how many variants a part has, we measure shortcuts by occlusion through $\lambda$ (occlusion also reaches the localizer): $\lambda$ averages 0.055 for the joint CBM, 0.036 for APN-lite, and $-0.004$ for \ocve{}, and this order holds in every split. With $i$ visible, the joint CBM reads $j$ as the co-occurring variant in 12--16\% of these cases, and \ocve{} does so in 0.1--0.3\%. In the main run, CEM's part accuracy likewise falls from 0.97--0.98 (seen) to 0.70--0.71 (unseen), and its $\lambda$ is 0.048.

\paragraph{Real images.} On unseen traffic-sign compositions in MTSD~\cite{ertler2020mtsd} (appendix), the joint CBM certifies wrong codes 99.8\% of the time (100\% with ViT-B/16 and design variants as separate pictograms). Of its wrong pictogram claims, 84\% name a pictogram that occurs with the sign's frame in seen codes, where pictogram frequencies predict 18\%. Isolated readers lower this share to 32\% (13\% expected) and certified risk to 67\% (81\% with color input); most remaining errors are in the pictogram. In two post hoc analyses (appendix), these readers' pictogram claims err far less often for frequent pictograms (26\% wrong above 259 training signs vs.\ 67--77\% for rarer ones), and a circular content window without the border lowers risk to 47\% (ViT-B/16: 44\%), which is still far too high.

\subsection{Stress tests: seen, not inferred}\label{sec:stress}
\paragraph{Decision-critical occlusion.} We select unseen-class images whose visible parts identify the class and gray out a part whose removal makes the visible evidence ambiguous; any single-class commitment must then rely on parts that are not visible. The selection uses only ground-truth masks (mean 782 images per split). Models without a certificate commit when their confidence exceeds the threshold that gives 1\% risk on seen validation images. Commitment rates are 2.3\% for \ocve{}, 15.5\% for APN-lite (11--15 points above \ocve{} per split), 17\% for B2-maskpool, and 31\% by confidence and 60\% by certificate for the joint CBM.
\paragraph{Evidence distance per image.} No certificate with $d(x)\ge2$ was wrong (0 of 1{,}195, seen and unseen pooled), against 22 of 3{,}529 at $d(x)=1$. Errors are few, however, and images with $d(x)\ge2$ also show more parts. A post hoc one-sided Fisher test gives $p=0.0016$, and stratifying by the number of truly visible parts gives $p=0.010$. Stratifying by the number of claimed parts, a model output that also bounds $d(x)$, gives $p=0.11$, as nearly all certificates with $d(x)\ge2$ claim four or five parts. This pattern is consistent with the bound but does not prove it.

\subsection{What isolation means}\label{sec:ablate}
\begin{figure*}[t]
\begin{minipage}[t]{0.6\textwidth}\centering
\includegraphics[width=\linewidth]{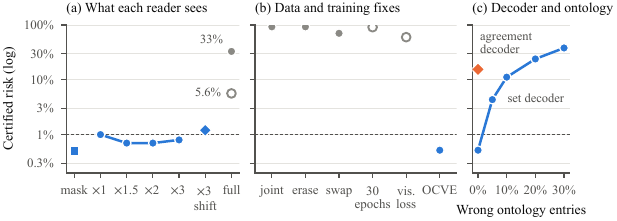}
\caption{What makes certificates trustworthy (FunnyBirds, unseen classes; certified risk on a log scale; dashed line: 1\%). (a) Any local window around the part keeps risk near 1\%; only the full image breaks it (A1). (b) Part erasure, part swap, training for 30 epochs, and a concept loss on visible parts only (vis.\ loss) do not repair the joint CBM (A2). (c) On \ocve{}'s own part claims, a decoder that only counts agreeing parts raises risk from 0.5\% to 15\%, and wrong ontology entries raise it steadily. Hollow: retrained with the localizer gate (Tables~\ref{tab:unseen} and~\ref{tab:ablate}).}
\label{fig:isolation}
\end{minipage}\hfill
\begin{minipage}[t]{0.38\textwidth}\centering
\includegraphics[width=\linewidth]{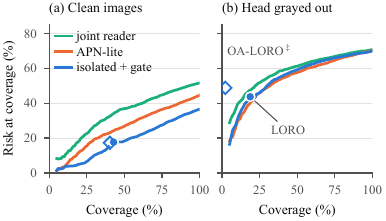}
\caption{CUB run 4: risk--coverage on unseen species (class vectors, mean of 3 splits). Isolated readers are ranked by LORO, then margin; the joint reader is ranked by its own LORO certificate; and APN-lite is ranked by confidence. Dot: LORO certificate. $^\ddagger$Diamond: occlusion-aware LORO (OA-LORO, post hoc).}
\label{fig:cub}
\end{minipage}
\end{figure*}

\begin{table}[t]\centering\small
\caption{Ablations on unseen FunnyBirds classes (3 splits). A1 varies the context each reader sees, including a crop whose center is shifted off the part; A2 tries to fix the joint CBM with data or training; A3 removes components of \ocve{} at inference. Rows come from the second training run (Sec.~\ref{sec:protocol}), except the full-image row, which is B2-ind of Table~\ref{tab:unseen}, and the row marked $^\ast$, which retrains the joint CBM and uses the localizer gate; Table~\ref{tab:unseen} gives both models retrained on visible parts. --: not computed.}
\label{tab:ablate}
\setlength{\tabcolsep}{3pt}
\begin{tabular}{llcc}
\toprule
& Variant & Cert.\ risk$\downarrow$ (cov.) & GZSL$\uparrow$\\
\midrule
\multirow{8}{*}{A1} & masked crop (\ocve{}) & 0.5\% (52\%) & 0.677\\
& one network, five heads & 0.9\% (52\%) & --\\
& box $\times1$ & 1.0\% (51\%) & 0.676\\
& box $\times1.5$ & 0.7\% (52\%) & 0.678\\
& box $\times2$ & 0.7\% (52\%) & 0.677\\
& box $\times3$ & 0.8\% (51\%) & 0.672\\
& box $\times3$, shifted center & 1.2\% (51\%) & 0.670\\
& full image (B2-ind) & 33.0\% (32\%) & 0.512\\
\midrule
\multirow{3}{*}{A2} & B2 + part erasure & 92.8\% (43\%) & 0.093\\
& B2 + part swap & 70.6\% (35\%) & 0.331\\
& B2, 30 epochs$^\ast$ & 91.3\% (38\%) & 0.145\\
\midrule
\multirow{3}{*}{A3} & no \textsc{absent} class & 0.5\% (52\%) & --\\
& no confidence threshold $\tau$ & 1.0\% (53\%) & --\\
& ground-truth masks at test & 4.9\% (46\%) & --\\
\bottomrule
\end{tabular}
\end{table}
\paragraph{Local windows.} A1 gives each reader a square crop centered on its part, 1--3 times the part's extent and without graying (Table~\ref{tab:ablate}, Fig.~\ref{fig:isolation}a). Certified risk stays at or below 1\% for every centered local crop and reaches 33\% only when the reader sees the whole image, although a $\times3$ crop contains 2.7 other parts on average. Shifting the crop center at random moves risk from 0.8\% to 1.2\%. In two post hoc controls, one network reading the masked crops of all five parts, with one head per part, certifies at 0.9\% (separate readers: 0.5\%), and a $\times1$ window at native scale and position certifies at 0.9\%, so magnification does not explain the low risk. The safe variants thus share a local window around the part, read in its own forward pass. Graying the tail, the part most correlated with the wing and present in 99\% of $\times3$ wing crops, changes the share of wing readings that match the tail's co-occurring variant by 0.1 points, against 4.0 points for the joint CBM (post hoc; appendix). A reader trained on a window around its part may learn that part's appearance rather than its context. B2-maskpool and B2-ind also err with small average shortcuts ($\lambda\le0.02$ vs.\ 0.055 for the joint CBM). For the visible-part B2-ind, the Grad-CAM peak lies on the part in 30\% of wrong and 86\% of correct readings, visible tails of unseen classes are read correctly in 73\% of cases, and $\times3$ windows at test still give 4.5\% risk (post hoc; appendix).
\paragraph{Data and training fixes.} Randomly erasing parts during training (loss masked for erased parts) or swapping a part for one cut from another class leaves the joint CBM's certified risk at 93\% and 71\%, more than 100 times \ocve{}'s (Fig.~\ref{fig:isolation}b). Retrained and scored with the gate, it certifies at 82\% risk; when its concept loss also covers only visible parts, it certifies at 60\%, with $\lambda$ at 0.034--0.045. B2-ind, CEM, and APN-lite treated the same way certify at 5.6\%, 72\%, and 40\% (Table~\ref{tab:unseen}, lower block; with $\tau$ for APN-lite). Visible-part training and gating thus remove most of B2-ind's risk but only part of the others'.
\paragraph{Components.} The \textsc{absent} class is redundant once $\tau$ is applied, so in \ocve{} $\bot$ comes from the localizer and $\tau$; removing $\tau$ doubles risk (0.5\% to 1.0\%). Feeding ground-truth masks to readers trained on predicted masks raises risk to 4.9\%, mostly by misreading small parts (post hoc; appendix), so training and test need the same localizer.
\paragraph{Decoder and ontology.} Two inference-only tests on \ocve{}'s stored claims separate the decoder and ontology from the readers (Fig.~\ref{fig:isolation}c). A rule that certifies when at least two claimed parts agree with the overall decision, without checking rivals, raises risk from 0.5\% to 15.4\% (at 79\% coverage) on the same, nearly perfect, claims. Restricted to the certificates with the most agreeing parts, up to \ocve{}'s coverage (post hoc), the rule still reaches 3.7\%, 7 times \ocve{}'s risk. Replacing a fraction of ontology entries with a wrong variant raises risk to 4.4, 11.2, 24, and 38\% at 5, 10, 20, and 30\% wrong entries. Set decoding does not guarantee fail-safe abstention when the ontology is wrong.
\paragraph{Backbones.} With ResNet-50, ViT-B/16, and the fine-tuned CLIP ViT-B/16 image encoder~\cite{radford2021clip} as backbones (frozen CLIP: appendix), certified risk is 0.42--0.57\% for \ocve{}, 4.6--6.6\% for B2-maskpool (ResNet-18: 13.8\%), and 89--96\% for the joint CBM. The order holds in all nine backbone--split pairs. \ocve{} commits on 2.2--4.0\% of decision-critical occlusions (APN-lite: 17--37\%), and $\lambda$ is unchanged (joint CBM 0.042--0.050, \ocve{} below 0).

\subsection{Seen classes and the oracle trap}\label{sec:seen}
Under nine corruptions of seen FunnyBirds classes (appendix), \ocve{} with oracle part masks appeared far more robust than a black box (risk 0.0\% vs.\ 21.2\% at the coverage of its $d(x)\ge2$ certificates). With a learned localizer the advantage reversed (19.7\% vs.\ 2.3\%); after matching training conditions (readers trained on predicted-mask crops, shared corruption augmentation for all models), the two tie in risk (0.0\% vs.\ 0.1\%) and error ranking (AUROC 0.952 vs.\ 0.944), although \ocve{}'s clean accuracy is 6 points lower (0.671 vs.\ 0.735). A class's code distance to its nearest rival does not predict its risk (Spearman $+0.14$). On seen classes the ontology adds no information the data lacks, and robustness gains measured with oracle masks did not carry over to a learned localizer. On the 34\% of the study's test images that lack parts FunnyBirds removed~\cite{hesse2023funnybirds}, even perfect part readers identify a unique class in only 30\% of cases (97\% of complete birds).

\subsection{Real images: CUB}\label{sec:cub}
\begin{table}[t]\centering\small
\caption{CUB-200-2011, 50 unseen species, mean over 3 splits. Run 4 is the final design, and KP repeats it with competitors given the same keypoints; development runs 1--3 are in the appendix. ZSL: per-class accuracy. Sup.: image-level (img) or class-level (cls) attributes. Risk: LORO certificates of the isolated readers and the other models at the same coverage (43\%; APN-lite in run 4: at the cls coverage). Occl.: ZSL with the head grayed out. $\bot$: separate visibility gate. Shading: best value per block.}
\label{tab:cub}
\setlength{\tabcolsep}{2.5pt}
\begin{tabular}{cllccc}
\toprule
Run & Model & Sup. & ZSL$\uparrow$ & Risk$\downarrow$ & Occl.$\uparrow$\\
\midrule
\multirow{5}{*}{4} & Joint attribute reader & img & 0.399 & 43.9\% & 0.223\\
& Joint attribute reader & cls & 0.479 & 34.3\% & 0.288\\
& Isolated + $\bot$ gate & img & 0.573 & 20.8\% & 0.247\\
& Isolated + $\bot$ gate & cls & \best{0.631} & \best{17.7\%} & \best{0.298}\\
& APN-lite & cls & 0.553 & 24.0\% & 0.294\\
\midrule
\multirow{4}{*}{KP} & Isolated + $\bot$ gate & cls & \best{0.627} & \best{17.5\%} & \best{0.304}\\
& Pooled in the same boxes & cls & 0.533 & 28.9\% & 0.273\\
& APN-lite & cls & 0.564 & 25.1\% & 0.298\\
& KP-APN & cls & 0.544 & 25.0\% & 0.276\\
\bottomrule
\end{tabular}
\end{table}
CUB has continuous attributes and no discrete code, so we certify with a stability rule. The attributes are divided among five body regions (head, wing, body, tail, leg), each read from its own crop around predicted keypoints, and a visibility gate per region drops the regions it judges hidden. A decision is certified when at least two regions remain and removing any one of them leaves the decision unchanged (leave-one-region-out, LORO), a continuous analog of tolerance $k=1$ that tests stability but not rival exclusion (examined below). Because scores are continuous, the bound of Sec.~\ref{sec:framework} does not apply, and CUB tests the readers. As in ZSL, decisions and certificates consider only the 50 unseen species, except in the 200-species evaluation below.

\paragraph{Final design (run 4).} With class-level attributes, the isolated readers reach 0.631 ZSL accuracy, against 0.553 for APN-lite and 0.479 for the joint reader, and lead APN-lite in every split and on the proposed split~\cite{xian2019zsl} (0.622 vs.\ 0.581). On clean images, 17.7\% of their LORO certificates are wrong, compared with 24.0\% for APN-lite and 34.3\% for the joint reader at the same 43\% coverage (95\% intervals of the differences over species: $-0.1$ to 13.5 points, not significant, and 12.1 to 21.5 points). Ranking the certified images by margin lowers their risk to 6.6\% at 25\% coverage, where APN-lite reaches 15.4\% (Fig.~\ref{fig:cub}a). Training is not deterministic on our hardware; over all trainings here and with ResNet-50, LORO risk stays below APN-lite's in every split (mean 0.60--0.83 times; appendix). Over six splits with three training seeds per model (post hoc), APN-lite's risk is 6.3 points higher (1.1 to 11.5).

\paragraph{Matched keypoint supervision.} Because APN-lite does not use part keypoints, block KP repeats run 4 with two competitors that do. The first pools the features of one shared ResNet-18 in the same boxes and uses the same gate and LORO certificate. KP-APN adds to APN-lite a loss pulling each attribute's prototype map toward the keypoints of its region (weight 1, not tuned). The isolated readers lead both in accuracy in every split (by 9.4 and 8.3 points on average), with LORO risk 0.61 and 0.70 times theirs at the same coverage. Pooling also changes parameter sharing and image resolution, but one network reading all five crops, or readers on crops of the bird image at pooling's $224^2$ resolution, match the separate readers' risk (18.6\% vs.\ 28.1\% for pooling; appendix). On three further splits not used during development, the accuracy lead holds (0.605 vs.\ 0.506 and 0.514), the mean risk ratios are 0.69 and 0.75, and pooling is 9.7 points riskier (interval 6.7 to 13.0).

\paragraph{Where isolation does not help.} With image-level labels, LORO risk is close to APN-lite's at equal coverage (20.8\% vs.\ 21.6\%). Over all 200 species at zero calibration offset, the isolated readers trail APN-lite on unseen species (0.251 vs.\ 0.313) yet have the larger area under the seen--unseen curve (AUSUC~\cite{chao2016empirical}: 0.427 vs.\ 0.381), consistent with an offset between seen and unseen scores. Their uncalibrated LORO certificates are wrong on 74\% of the unseen images they cover. With calibrated stacking tuned on the test images for both models, they reach the higher harmonic mean (0.569 vs.\ 0.523) but certify pooled seen and unseen images at a risk similar to APN-lite's (25.5\% vs.\ 24.8\%).

With the head grayed out, the gate reports it as unseen in 82\% of images (1.1\% on clean images) and lowers LORO risk on them by 7.1 points, yet APN-lite's confidence ranks errors better (39.9\% vs.\ 43.7\%; Fig.~\ref{fig:cub}b). On SUB~\cite{bader2025sub}, generated images of known species with one attribute replaced, the isolated readers rank the substituted value above the typical one on 55.9\% of images (joint reader 44.4\%, APN-lite 60.1\%), but APN-lite also does so on 31.6\% of real images (isolated readers: 13.9\%). By region, isolated readers trail APN-lite only on the head (appendix).

\paragraph{Rival exclusion and the decoder (post hoc).} Occlusion-aware LORO also requires that no rival wins when the dropped regions are filled with that rival's attributes, the analog of set decoding. On clean images it certifies 40.3\% of images at 17.3\% risk (APN-lite: 23.4\% at that coverage). With the head grayed out, it abstains on 98\% of images, whereas plain LORO is wrong on 41\% of them even when the gate reports the head as unseen. APN-lite's confidence still finds a small set it gets right (6.3\% risk at 2.2\% coverage). Unlike on FunnyBirds, region agreement alone reaches 19.4\% risk at LORO's coverage (LORO: 17.7\%; appendix), so the decoder matters little.

\paragraph{Comparability.} All CUB models use ground-truth bounding-box crops, and our splits differ from the standard benchmark, so accuracies are not comparable with published numbers. Under the original APN protocol (ResNet-101, 448\,px full images, proposed split), APN-lite reaches 63.1\% ZSL accuracy, 8.9 points below APN's 72.0\%~\cite{xu2020apn}, so APN-lite is a simplified baseline; the comparison with pooled features does not depend on it.

\section{Discussion and limitations}
In our experiments, the ontology helped only where it carried information the training data lacked, and not on seen classes. On unseen FunnyBirds classes, certificates had low risk only with local reading, a rival-excluding decoder, and a correct ontology. On clean CUB images, isolated readers were more accurate than an attribute-prototype model and less risky than both this model (by about 6 points) and pooled features.

\paragraph{Limitations.} The decisive evidence on shortcuts comes from a synthetic benchmark; on real traffic signs, 67\% of the isolated readers' certificates are wrong (Sec.~\ref{sec:mechanism}). Local windows reduced certified risk in our settings; this does not show that they are necessary or sufficient in general. Full-image part readers trained on visible parts show a small average shortcut yet remain about ten times riskier, and their errors persist with local windows at test; why training on the full image produces them is not established (Sec.~\ref{sec:ablate}). Our readers need part masks or keypoints for training (competitors given them still certified with higher risk). Isolation depends on a localizer that sees the whole image (leakage: appendix) and separates regions but not the concepts within one (\eg, the 112 attributes of a CUB head reader), so $\lambda$ does not capture shortcuts within a region. Hard isolation also costs seen-class accuracy (Table~\ref{tab:unseen}: 0.605 for \ocve{} vs.\ 0.696--0.730 for models that predict every part), mainly through the localizer and readers (appendix). Spatial and prototype-grounded CBMs~\cite{panousis2024coarse,ramachandram2026pfcbm,colamonaco2026pgcm} were tested only through proxies. We did not test VLM concept readers or run a user study, and several analyses are post hoc.

\section*{Data and ethics}
All datasets are public. FunnyBirds~\cite{hesse2023funnybirds} is a synthetic dataset distributed by its authors. CUB-200-2011~\cite{wah2011cub} contains photographs whose use its providers restrict to non-commercial research and education. MTSD~\cite{ertler2020mtsd} is released under CC BY-NC-SA, and SUB~\cite{bader2025sub} is released under CC BY-NC 4.0. We use all four datasets for non-commercial research only, and the paper shows no MTSD or SUB images. The study collects no new data from people.
{
\small
\bibliographystyle{ieeenat_fullname}
\bibliography{main}

\begin{thebibliography}{48}
\providecommand{\natexlab}[1]{#1}
\providecommand{\url}[1]{\texttt{#1}}
\expandafter\ifx\csname urlstyle\endcsname\relax
  \providecommand{\doi}[1]{doi: #1}\else
  \providecommand{\doi}{doi: \begingroup \urlstyle{rm}\Url}\fi

\bibitem[Bader et~al.(2025)Bader, Girrbach, Alaniz, and Akata]{bader2025sub}
Jessica Bader, Leander Girrbach, Stephan Alaniz, and Zeynep Akata.
\newblock {SUB}: Benchmarking {CBM} generalization via synthetic attribute
  substitutions.
\newblock In \emph{ICCV}, 2025.

\bibitem[Benou and Riklin-Raviv(2025)]{benou2025salf}
Itay Benou and Tammy Riklin-Raviv.
\newblock Show and tell: Visually explainable deep neural nets via
  spatially-aware concept bottleneck models.
\newblock In \emph{CVPR}, 2025.

\bibitem[Berge(1989)]{berge1989hypergraphs}
Claude Berge.
\newblock \emph{Hypergraphs: Combinatorics of Finite Sets}.
\newblock North-Holland, 1989.

\bibitem[Branson et~al.(2014)Branson, Van~Horn, Belongie, and
  Perona]{branson2014pose}
Steve Branson, Grant Van~Horn, Serge Belongie, and Pietro Perona.
\newblock Bird species categorization using pose normalized deep convolutional
  nets.
\newblock In \emph{BMVC}, 2014.

\bibitem[Chao et~al.(2016)Chao, Changpinyo, Gong, and Sha]{chao2016empirical}
Wei-Lun Chao, Soravit Changpinyo, Boqing Gong, and Fei Sha.
\newblock An empirical study and analysis of generalized zero-shot learning for
  object recognition in the wild.
\newblock In \emph{ECCV}, 2016.

\bibitem[Chen et~al.(2022)Chen, Hong, Liu, Xie, Sun, Li, Peng, Lu, and
  You]{chen2022transzero}
Shiming Chen, Ziming Hong, Yang Liu, Guo-Sen Xie, Baigui Sun, Hao Li, Qinmu
  Peng, Ke Lu, and Xinge You.
\newblock {TransZero}: Attribute-guided transformer for zero-shot learning.
\newblock In \emph{AAAI}, 2022.

\bibitem[Colamonaco et~al.(2026)Colamonaco, Debot, Barbiero, and
  Marra]{colamonaco2026pgcm}
Stefano Colamonaco, David Debot, Pietro Barbiero, and Giuseppe Marra.
\newblock Prototype-grounded concept models for verifiable concept alignment.
\newblock In \emph{ICML}, 2026.

\bibitem[Dietterich and Bakiri(1995)]{dietterich1995ecoc}
Thomas~G. Dietterich and Ghulum Bakiri.
\newblock Solving multiclass learning problems via error-correcting output
  codes.
\newblock \emph{Journal of Artificial Intelligence Research}, 2:\penalty0
  263--286, 1995.

\bibitem[Dosovitskiy et~al.(2021)Dosovitskiy, Beyer, Kolesnikov, Weissenborn,
  Zhai, Unterthiner, Dehghani, Minderer, Heigold, Gelly, Uszkoreit, and
  Houlsby]{dosovitskiy2021vit}
Alexey Dosovitskiy, Lucas Beyer, Alexander Kolesnikov, Dirk Weissenborn,
  Xiaohua Zhai, Thomas Unterthiner, Mostafa Dehghani, Matthias Minderer, Georg
  Heigold, Sylvain Gelly, Jakob Uszkoreit, and Neil Houlsby.
\newblock An image is worth 16x16 words: Transformers for image recognition at
  scale.
\newblock In \emph{ICLR}, 2021.

\bibitem[Elhoseiny et~al.(2017)Elhoseiny, Zhu, Zhang, and
  Elgammal]{elhoseiny2017link}
Mohamed Elhoseiny, Yizhe Zhu, Han Zhang, and Ahmed Elgammal.
\newblock Link the head to the ``beak'': Zero shot learning from noisy text
  description at part precision.
\newblock In \emph{CVPR}, 2017.

\bibitem[Ertler et~al.(2020)Ertler, Mislej, Ollmann, Porzi, Neuhold, and
  Kuang]{ertler2020mtsd}
Christian Ertler, Jerneja Mislej, Tobias Ollmann, Lorenzo Porzi, Gerhard
  Neuhold, and Yubin Kuang.
\newblock The {Mapillary} traffic sign dataset for detection and classification
  on a global scale.
\newblock In \emph{ECCV}, 2020.

\bibitem[Espinosa~Zarlenga(2026)]{espinosa2026defense}
Mateo Espinosa~Zarlenga.
\newblock In defense of information leakage in concept-based models.
\newblock In \emph{ICML (Position Paper Track)}, 2026.

\bibitem[Espinosa~Zarlenga et~al.(2022)Espinosa~Zarlenga, Barbiero, Ciravegna,
  Marra, Giannini, Diligenti, Shams, Precioso, Melacci, Weller, Li{\`o}, and
  Jamnik]{espinosa2022cem}
Mateo Espinosa~Zarlenga, Pietro Barbiero, Gabriele Ciravegna, Giuseppe Marra,
  Francesco Giannini, Michelangelo Diligenti, Zohreh Shams, Frederic Precioso,
  Stefano Melacci, Adrian Weller, Pietro Li{\`o}, and Mateja Jamnik.
\newblock Concept embedding models: Beyond the accuracy-explainability
  trade-off.
\newblock In \emph{NeurIPS}, 2022.

\bibitem[Geifman and El-Yaniv(2017)]{geifman2017selective}
Yonatan Geifman and Ran El-Yaniv.
\newblock Selective classification for deep neural networks.
\newblock In \emph{NeurIPS}, 2017.

\bibitem[Havasi et~al.(2022)Havasi, Parbhoo, and
  Doshi-Velez]{havasi2022leakage}
Marton Havasi, Sonali Parbhoo, and Finale Doshi-Velez.
\newblock Addressing leakage in concept bottleneck models.
\newblock In \emph{NeurIPS}, 2022.

\bibitem[He et~al.(2016)He, Zhang, Ren, and Sun]{he2016resnet}
Kaiming He, Xiangyu Zhang, Shaoqing Ren, and Jian Sun.
\newblock Deep residual learning for image recognition.
\newblock In \emph{CVPR}, 2016.

\bibitem[Heidemann et~al.(2023)Heidemann, Monnet, and
  Roscher]{heidemann2023correlation}
Lena Heidemann, Maureen Monnet, and Karsten Roscher.
\newblock Concept correlation and its effects on concept-based models.
\newblock In \emph{WACV}, 2023.

\bibitem[Hesse et~al.(2023)Hesse, Schaub-Meyer, and Roth]{hesse2023funnybirds}
Robin Hesse, Simone Schaub-Meyer, and Stefan Roth.
\newblock {FunnyBirds}: A synthetic vision dataset for a part-based analysis of
  explainable {AI} methods.
\newblock In \emph{ICCV}, 2023.

\bibitem[Honkala and Laihonen(2007)]{honkala2007robust}
Iiro Honkala and Tero Laihonen.
\newblock On identifying codes that are robust against edge changes.
\newblock \emph{Information and Computation}, 205:\penalty0 1078--1095, 2007.

\bibitem[Huang et~al.(2016)Huang, Xu, Tao, and Zhang]{huang2016partstacked}
Shaoli Huang, Zhe Xu, Dacheng Tao, and Ya Zhang.
\newblock Part-stacked {CNN} for fine-grained visual categorization.
\newblock In \emph{CVPR}, 2016.

\bibitem[Huynh and Elhamifar(2020)]{huynh2020dazle}
Dat Huynh and Ehsan Elhamifar.
\newblock Fine-grained generalized zero-shot learning via dense attribute-based
  attention.
\newblock In \emph{CVPR}, 2020.

\bibitem[Ignatiev et~al.(2020)Ignatiev, Narodytska, Asher, and
  Marques-Silva]{ignatiev2020contrastive}
Alexey Ignatiev, Nina Narodytska, Nicholas Asher, and Joao Marques-Silva.
\newblock From contrastive to abductive explanations and back again.
\newblock In \emph{AIxIA}, 2020.

\bibitem[Izza et~al.(2024)Izza, Huang, Morgado, Planes, Ignatiev, and
  Marques-Silva]{izza2024distance}
Yacine Izza, Xuanxiang Huang, Antonio Morgado, Jordi Planes, Alexey Ignatiev,
  and Joao Marques-Silva.
\newblock Distance-restricted explanations: Theoretical underpinnings \&
  efficient implementation.
\newblock In \emph{KR}, 2024.

\bibitem[Jayaraman et~al.(2014)Jayaraman, Sha, and
  Grauman]{jayaraman2014decorrelating}
Dinesh Jayaraman, Fei Sha, and Kristen Grauman.
\newblock Decorrelating semantic visual attributes by resisting the urge to
  share.
\newblock In \emph{CVPR}, 2014.

\bibitem[Joren et~al.(2024)Joren, Marx, and Ustun]{joren2024safeguards}
Hailey Joren, Charles Marx, and Berk Ustun.
\newblock Classification with conceptual safeguards.
\newblock In \emph{ICLR}, 2024.

\bibitem[Karpovsky et~al.(1998)Karpovsky, Chakrabarty, and
  Levitin]{karpovsky1998identifying}
Mark~G. Karpovsky, Krishnendu Chakrabarty, and Lev~B. Levitin.
\newblock On a new class of codes for identifying vertices in graphs.
\newblock \emph{IEEE Transactions on Information Theory}, 44\penalty0
  (2):\penalty0 599--611, 1998.

\bibitem[Kim et~al.(2023)Kim, Jung, Park, Kim, and Yoon]{kim2023probcbm}
Eunji Kim, Dahuin Jung, Sangha Park, Siwon Kim, and Sungroh Yoon.
\newblock Probabilistic concept bottleneck models.
\newblock In \emph{ICML}, 2023.

\bibitem[Kim et~al.(2022)Kim, Meister, Ramaswamy, Fong, and
  Russakovsky]{kim2022hive}
Sunnie S.~Y. Kim, Nicole Meister, Vikram~V. Ramaswamy, Ruth Fong, and Olga
  Russakovsky.
\newblock {HIVE}: Evaluating the human interpretability of visual explanations.
\newblock In \emph{ECCV}, 2022.

\bibitem[Koh et~al.(2020)Koh, Nguyen, Tang, Mussmann, Pierson, Kim, and
  Liang]{koh2020cbm}
Pang~Wei Koh, Thao Nguyen, Yew~Siang Tang, Stephen Mussmann, Emma Pierson, Been
  Kim, and Percy Liang.
\newblock Concept bottleneck models.
\newblock In \emph{ICML}, 2020.

\bibitem[La~Malfa et~al.(2021)La~Malfa, Michelmore, Zbrzezny, Paoletti, and
  Kwiatkowska]{lamalfa2021robust}
Emanuele La~Malfa, Rhiannon Michelmore, Agnieszka~M. Zbrzezny, Nicola Paoletti,
  and Marta Kwiatkowska.
\newblock On guaranteed optimal robust explanations for {NLP} models.
\newblock In \emph{IJCAI}, 2021.

\bibitem[Lampert et~al.(2009)Lampert, Nickisch, and Harmeling]{lampert2009dap}
Christoph~H. Lampert, Hannes Nickisch, and Stefan Harmeling.
\newblock Learning to detect unseen object classes by between-class attribute
  transfer.
\newblock In \emph{CVPR}, 2009.

\bibitem[Mahinpei et~al.(2021)Mahinpei, Clark, Lage, Doshi-Velez, and
  Pan]{mahinpei2021promises}
Anita Mahinpei, Justin Clark, Isaac Lage, Finale Doshi-Velez, and Weiwei Pan.
\newblock Promises and pitfalls of black-box concept learning models.
\newblock \emph{arXiv:2106.13314}, 2021.

\bibitem[Mancini et~al.(2021)Mancini, Naeem, Xian, and Akata]{mancini2021open}
Massimiliano Mancini, Muhammad~Ferjad Naeem, Yongqin Xian, and Zeynep Akata.
\newblock Open world compositional zero-shot learning.
\newblock In \emph{CVPR}, 2021.

\bibitem[Marconato et~al.(2023)Marconato, Teso, Vergari, and
  Passerini]{marconato2023reasoning}
Emanuele Marconato, Stefano Teso, Antonio Vergari, and Andrea Passerini.
\newblock Not all neuro-symbolic concepts are created equal: Analysis and
  mitigation of reasoning shortcuts.
\newblock In \emph{NeurIPS}, 2023.

\bibitem[Nayak et~al.(2023)Nayak, Yu, and Bach]{nayak2023csp}
Nihal~V. Nayak, Peilin Yu, and Stephen~H. Bach.
\newblock Learning to compose soft prompts for compositional zero-shot
  learning.
\newblock In \emph{ICLR}, 2023.

\bibitem[Nguyen et~al.(2021)Nguyen, Kim, and Nguyen]{nguyen2021effectiveness}
Giang Nguyen, Daeyoung Kim, and Anh Nguyen.
\newblock The effectiveness of feature attribution methods and its correlation
  with automatic evaluation scores.
\newblock In \emph{NeurIPS}, 2021.

\bibitem[Palatucci et~al.(2009)Palatucci, Pomerleau, Hinton, and
  Mitchell]{palatucci2009zero}
Mark Palatucci, Dean Pomerleau, Geoffrey~E. Hinton, and Tom~M. Mitchell.
\newblock Zero-shot learning with semantic output codes.
\newblock In \emph{NeurIPS}, 2009.

\bibitem[Panousis et~al.(2024)Panousis, Ienco, and Marcos]{panousis2024coarse}
Konstantinos~P. Panousis, Dino Ienco, and Diego Marcos.
\newblock Coarse-to-fine concept bottleneck models.
\newblock In \emph{NeurIPS}, 2024.

\bibitem[Parisini et~al.(2025)Parisini, Chakraborti, Harbron, MacArthur, and
  Banerji]{parisini2025leakage}
Enrico Parisini, Tapabrata Chakraborti, Chris Harbron, Ben~D. MacArthur, and
  Christopher R.~S. Banerji.
\newblock Leakage and interpretability in concept-based models.
\newblock \emph{arXiv:2504.14094}, 2025.

\bibitem[Radford et~al.(2021)Radford, Kim, Hallacy, Ramesh, Goh, Agarwal,
  Sastry, Askell, Mishkin, Clark, Krueger, and Sutskever]{radford2021clip}
Alec Radford, Jong~Wook Kim, Chris Hallacy, Aditya Ramesh, Gabriel Goh,
  Sandhini Agarwal, Girish Sastry, Amanda Askell, Pamela Mishkin, Jack Clark,
  Gretchen Krueger, and Ilya Sutskever.
\newblock Learning transferable visual models from natural language
  supervision.
\newblock In \emph{ICML}, 2021.

\bibitem[Ramachandram(2026)]{ramachandram2026pfcbm}
Dhanesh Ramachandram.
\newblock Spatially grounded concept bottleneck models via part-factorized
  attention.
\newblock \emph{arXiv:2606.04364}, 2026.

\bibitem[Raman et~al.(2025)Raman, Espinosa~Zarlenga, Heo, and
  Jamnik]{raman2025locality}
Naveen Raman, Mateo Espinosa~Zarlenga, Juyeon Heo, and Mateja Jamnik.
\newblock Do concept bottleneck models respect localities?
\newblock \emph{TMLR}, 2025.

\bibitem[Saralajew et~al.(2026)Saralajew, Poeta, Pastor, and
  Shaker]{saralajew2026auditing}
Sascha Saralajew, Eleonora Poeta, Eliana Pastor, and Ammar Shaker.
\newblock Auditing where concept bottleneck accuracy comes from on {CUB}.
\newblock \emph{arXiv:2610.28059}, 2026.

\bibitem[Wah et~al.(2011)Wah, Branson, Welinder, Perona, and
  Belongie]{wah2011cub}
Catherine Wah, Steve Branson, Peter Welinder, Pietro Perona, and Serge
  Belongie.
\newblock The {Caltech-UCSD} birds-200-2011 dataset.
\newblock Technical Report CNS-TR-2011-001, California Institute of Technology,
  2011.

\bibitem[Wang et~al.(2021)Wang, Khosravi, and Van~den
  Broeck]{wang2021probabilistic}
Eric Wang, Pasha Khosravi, and Guy Van~den Broeck.
\newblock Probabilistic sufficient explanations.
\newblock In \emph{IJCAI}, 2021.

\bibitem[Xian et~al.(2019)Xian, Lampert, Schiele, and Akata]{xian2019zsl}
Yongqin Xian, Christoph~H. Lampert, Bernt Schiele, and Zeynep Akata.
\newblock Zero-shot learning -- a comprehensive evaluation of the good, the bad
  and the ugly.
\newblock \emph{TPAMI}, 41\penalty0 (9):\penalty0 2251--2265, 2019.

\bibitem[Xu et~al.(2020)Xu, Xian, Wang, Schiele, and Akata]{xu2020apn}
Wenjia Xu, Yongqin Xian, Jiuniu Wang, Bernt Schiele, and Zeynep Akata.
\newblock Attribute prototype network for zero-shot learning.
\newblock In \emph{NeurIPS}, 2020.

\bibitem[Zhang et~al.(2014)Zhang, Donahue, Girshick, and
  Darrell]{zhang2014partrcnn}
Ning Zhang, Jeff Donahue, Ross Girshick, and Trevor Darrell.
\newblock Part-based {R-CNN}s for fine-grained category detection.
\newblock In \emph{ECCV}, 2014.

\end{thebibliography}
}
\clearpage
\appendix
\setcounter{table}{0}
\setcounter{figure}{0}
\renewcommand{\thetable}{S\arabic{table}}
\renewcommand{\thefigure}{S\arabic{figure}}
\renewcommand{\theHtable}{S\arabic{table}}
\renewcommand{\theHfigure}{S\arabic{figure}}
\paragraph{Overview.} Sec.~\ref{sec:s-impl} gives the implementation of every model and experiment. Sec.~\ref{sec:s-fb} extends the FunnyBirds results of Secs.~\ref{sec:unseen}--\ref{sec:ablate}: backbones, evidence distance by stratum, complete birds, retraining on visible parts, training length and compute, attribution maps, native-scale windows, confidence ranking, and calibrated stacking. Sec.~\ref{sec:s-cub} adds the CUB results of Sec.~\ref{sec:cub}: development runs 1--3, matched controls, all 200 species, training seeds, and SUB. Sec.~\ref{sec:s-mtsd} describes the traffic-sign study of Sec.~\ref{sec:mechanism}, and Sec.~\ref{sec:s-seen} reports the corruption study on seen classes (Sec.~\ref{sec:seen}). Tables and figures here are numbered S1, S2, and so on; other numbers refer to the main text.
\paragraph{Post hoc analyses.} Analyses decided after seeing a result are post hoc. They include the analyses marked post hoc in the main text or here, the later experiments (CLIP, SUB, visible-part retraining of four baselines, native-scale windows for \ocve{} and B2-ind, the shared-reader, resolution, and head sub-region controls, the tail-occlusion probe, the MTSD circular window and design variants, the CUB seed runs, and the 30-epoch joint CBM and CEM), and all chance levels, confidence intervals, independence-model references, attribution maps, error breakdowns, and 200-species evaluations. All four CUB runs are reported (Sec.~\ref{sec:cub} and Sec.~\ref{sec:s-cub}).

\section{Implementation details}\label{sec:s-impl}

\subsection{Common training setup}
All recognition models start from ImageNet weights of torchvision and take ImageNet-normalized inputs. They are trained in single precision with AdamW (weight decay $10^{-4}$) and a one-cycle learning-rate schedule with PyTorch's default shape, stepped every mini-batch. Unless stated otherwise, ResNet models use a learning rate of $3\cdot10^{-4}$ and batches of 128, and ViT-B/16 uses $10^{-4}$ and 64. Seeds are fixed per model and split. Unless stated otherwise, the only augmentation on FunnyBirds and CUB is a horizontal flip of the whole mini-batch with probability 0.5.

\subsection{FunnyBirds}\label{sec:s-impl-fb}
\paragraph{Data and splits.} Images are rendered at $256^2$. For each split seed, the unseen classes are chosen by walking a random permutation of the 50 classes and adding a class whenever every part variant still occurs among the remaining classes, until 10 classes are held out. The official training images of the 40 seen classes are shuffled and divided into 12{,}000 training images, 1{,}000 validation images, used only to set thresholds, and 1{,}000 seen-class test images. The unseen-class test set contains the 100 official test images and 2{,}000 random official training images of the unseen classes.

\paragraph{Localizer.} A ResNet-18 trunk feeds an FPN-style decoder: $1\times1$ lateral convolutions from the four stages to 96 channels, top-down upsampling and addition down to stride 4, a $3\times3$ convolution with batch normalization and ReLU, and a $1\times1$ convolution to six classes (background and the five part groups), upsampled to the input size. It is trained on the $256^2$ images for 5 epochs with class-weighted cross-entropy (weights proportional to the inverse square root of class frequency), learning rate $10^{-3}$, and batches of 32, flipping images and label maps together. A part group counts as present when its predicted mask (pixels whose argmax label is that group) has at least 15 pixels.

\paragraph{Readers and claims.} A reader's input is the bounding box of the predicted mask, padded by 4 pixels, with all pixels outside the mask set to gray (0.5), padded to a square with gray, and resized to $96^2$. Each group has its own ResNet-18 with $V_g+1$ outputs, where the extra output is \textsc{absent}; readers are trained for 5 epochs with cross-entropy. Training labels come from the localizer's predictions on the training images: the true variant if at least half of the predicted pixels belong to the group, and \textsc{absent} otherwise. The threshold $\tau_g$ is the 1\% quantile of the largest variant probability over correct claims on the validation images. A reader returns $\bot$ when \textsc{absent} has the highest probability or the top variant probability is below $\tau_g$, and the most probable variant otherwise.

\paragraph{Decoder.} A certificate is issued when exactly one of the 50 classes is compatible with all non-$\bot$ claims; no certificate is issued when all claims are $\bot$ or no class is compatible. The soft decoder sums $\log p_g$ of each candidate's codeword over the groups whose \textsc{absent} output does not win (a group without a predicted mask counts as \textsc{absent}). Here $p_g$ is the reader's softmax over its $V_g+1$ outputs, without renormalization, and the candidates are all 50 classes (GZSL) or the 10 unseen classes (ZSL).

\paragraph{Baselines.} All baselines see the full image resized to $224^2$ and are trained for 5 epochs; each uses its own thresholds, set by the same 1\% rule on correct validation predictions.
The joint CBM (B2) attaches five linear heads (no \textsc{absent} output) to one trunk and minimizes the sum of the five cross-entropies. B2-ind trains one ResNet-18 per group on the full image. The joint CBM, B2-ind, CEM, and APN-lite take the class's variant as the target for every part, including parts removed from the image or hidden by the viewpoint.
B2-maskpool takes layer3 features concatenated with upsampled layer4 features ($14\times14$, 768 channels for ResNet-18) and pools them for each group with weights equal to the fraction of each cell covered by the predicted mask, normalized to sum to one; each group has a linear head with an \textsc{absent} output and the same training labels as the readers.
APN-lite uses the same feature map, one prototype per attribute slot (26 one-hot variant slots), attribute scores equal to the spatial maximum of the prototype--feature dot product, and a global linear branch on pooled features. Its loss is the cross-entropy over seen classes of the compatibility between the global branch and the class codewords, plus a mean squared error between attribute scores and the class codeword, plus group-lasso and compactness terms weighted by $10^{-3}$; it uses batches of 64. Its confidence is the maximum softmax of the global compatibility over the candidate classes. For the prototype certificate of Table~\ref{tab:unseen}, a separately trained APN-lite with the same settings claims the highest-scoring prototype in every group, without a threshold, and these claims are decoded like \ocve{}'s.
CEM generates one 16-dimensional embedding per variant (linear layer and LeakyReLU), scores the embeddings of a group with a shared linear layer, and mixes them by their softmax probabilities; a linear task head over the 40 seen classes takes the five group embeddings. During training, each group's probabilities are replaced by the true one-hot vector with probability 0.25, independently per image and group. The loss is the task cross-entropy plus the sum of the concept cross-entropies.

\paragraph{Stress tests and shortcut measures.} For decision-critical occlusion, we keep unseen-class test images whose truly visible parts (ground-truth mask of at least 15 pixels) identify a unique class and that contain at least one visible part whose removal leaves two or more compatible classes; one such part is chosen at random, and its ground-truth pixels are set to gray before any model, including the localizer, sees the image. Models without a certificate commit when their confidence exceeds the smallest threshold that gives at most 1\% risk on the seen-class validation images. The shortcut rate counts wrong part predictions that equal the variant most often co-occurring, among seen classes, with the true variant of another part of the class. For $\lambda$, each part $i$ in turn is grayed out in all unseen-class test images; for every other part $j$, the target $\tilde a_j(a_i)$ is the variant of $j$ that co-occurs most with the true variant of $i$ among seen classes (ties go to the variant that occurs first in class order), and only images where both parts are visible and the target differs from the true variant of $j$ are used. $\lambda$ is the unweighted mean, over the ordered pairs with at least one such image, of $P(\hat a_j=\tilde a_j(a_i))$ on clean minus occluded images. Readings are argmax predictions for models without a localizer (joint CBM, CEM, B2-ind, and APN-lite) and claims for models with one (\ocve{}, B2-maskpool, and the window readers), so $\bot$ never counts as the target.

\paragraph{Figure examples.} The example of Fig.~\ref{fig:method} is the first unseen-class test image of split~0 (second training run), by index, with exactly one part removed, on which \ocve{} reports that part as $\bot$, reads the other four correctly, and abstains with a compatible set of two or three classes that contains the true class, while the joint CBM certifies a wrong class (56 of 2{,}100 images qualify).
\paragraph{Ablations and backbones.} The A1 crops are squares around the part's box enlarged by the stated factor and resized to $96^2$, $144^2$, $192^2$, and $224^2$ for factors 1, 1.5, 2, and 3; the shifted-center variant uses $\times3$ crops at $224^2$. The upper block of Table~\ref{tab:unseen} reports the main training run; ablations and stress tests use a second training run with the same settings, in which each experiment trains its own localizer. For the backbone checks, the readers, the joint CBM, B2-maskpool, and APN-lite are replaced by ResNet-50 (layer3 and layer4 features, 3{,}072 channels, for the feature-map models) or ViT-B/16 (inputs resized to $224^2$, $14\times14$ patch tokens); the localizer is the ResNet-18 segmenter of the second run on the same split, and B2-ind is not trained.

\subsection{CUB}\label{sec:s-impl-cub}
\paragraph{Data and splits.} A random split holds out the first 50 classes of a random permutation of the 200 classes; the proposed split follows~\cite{xian2019zsl}. Training uses the official training images of the 150 seen classes (about 4{,}495 images), the ZSL test set consists of the official test images of the unseen classes (1{,}437--1{,}468 images), and the AUSUC analysis also uses the official test images of the seen classes. Every model sees a square crop around the ground-truth bird box with side $1.15\max(w,h)$, resized to $320^2$; the keypoint model, the joint reader, APN-lite, and the pooling competitor see this crop resized to $224^2$. Unless stated otherwise, CUB models are trained for 8 epochs with batches of 64.

\paragraph{Keypoint model.} The keypoint model uses the same FPN design on a ResNet-18 trunk and outputs 15 heatmaps at stride 4 ($56^2$), plus a visibility head on pooled layer4 features. It is trained for 12 epochs (learning rate $10^{-3}$, no flips) with a binary cross-entropy on the heatmaps (positive weight 50; Gaussian targets with $\sigma=2$ cells, zero for invisible parts) and a binary cross-entropy on visibility. A keypoint is the heatmap argmax, and a part is visible when its visibility score exceeds a per-part threshold that maximizes F1 on the training images; no held-out images are used for this choice, and the pooling competitor uses the same keypoints.

\paragraph{Regions.} Table~\ref{tab:s-regions} lists the five regions. A region is present when any of its keypoints is predicted visible. The bird-box side is the longer side of the ground-truth bird box in the $320^2$ crop ($320/1.15$ pixels). With one visible keypoint, the region box is a square of 0.35 times this side around the point; with more, it is the bounding box of the points padded by 0.15 times this side. The box is then made square by extending its shorter side about its center, cropped from the $320^2$ image (pixels beyond the image border are black), and resized to $128^2$.

\begin{table}[!htb]\centering\small
\caption{CUB regions: parts that define each region and the attribute groups its reader predicts (278 of the 312 attributes; the 34 whole-bird attributes of primary color, shape, and size are not used).}
\label{tab:s-regions}
\setlength{\tabcolsep}{3pt}
\resizebox{\columnwidth}{!}{%
\begin{tabular}{llp{4.6cm}r}
\toprule
Region & Parts & Attribute groups & \#\\
\midrule
head & beak, crown, forehead, left and right eye, nape, throat & bill shape, bill length, bill color, head pattern, crown, forehead, nape, throat, and eye color & 112\\
wing & left and right wing & wing color, shape, pattern & 24\\
body & back, belly, breast & upperparts, underparts, back, breast, belly color; breast, back, belly pattern & 87\\
tail & tail & tail shape, tail pattern, upper and under tail color & 40\\
leg & left and right leg & leg color & 15\\
\bottomrule
\end{tabular}}
\end{table}

\paragraph{Gates and readers.} Each region has a ResNet-18 visibility gate on its $128^2$ crop, trained with a binary cross-entropy whose positive class is ``absent''. Its training crops are regions predicted present and truly visible (visible), regions predicted present but truly hidden (absent), and, for each image with probability 0.25, the crop of another present region of the same image (absent). At test time a region is kept when it is predicted present and the gate's absence probability is below 0.5. Each region reader is a ResNet-18 trained with a binary cross-entropy on every training image in which the keypoint model finds its region, whether or not the region is truly visible, against class-level attributes (class attribute percentages divided by 100) or image-level MTurk labels; with MTurk labels, attributes that have no annotation for the image are masked out of the loss. The joint reader is a ResNet-18 that predicts all 278 attributes from the full crop.

\paragraph{Scoring and certificates.} A class score is the cosine between the predicted attribute vector and the class vector, both centered by the mean seen-class vector and restricted to the attributes of the regions the gate keeps. ZSL and GZSL report per-class accuracy over the 50 unseen or all 200 classes. Certificates use the ZSL candidate set: LORO certifies the decision among the 50 unseen species when at least two regions are kept and removing the attributes of any single kept region leaves that decision unchanged. Occlusion-aware LORO (main text, post hoc) additionally fills the attributes of every region the gate dropped with each rival's class attributes and certifies only if no rival then wins. To compare risk at equal coverage, competitors keep their top-ranked images: APN-lite by its maximum softmax compatibility, and the joint reader and the pooling competitor by their own LORO certificate and then the margin between the two best classes.

\paragraph{APN-lite and the matched-keypoint competitors.} APN-lite on CUB uses 278 prototypes in 25 attribute groups, the losses described for FunnyBirds, and the global branch for prediction. Under the original APN protocol~\cite{xu2020apn}, it is trained on full images at $448^2$ with a ResNet-101, all 312 attributes, batches of 16, learning rate $10^{-4}$, and the proposed split. The pooling competitor (POOL) shares one ResNet-18 trunk, pools its $14\times14$, 768-channel features in each region box with weights equal to the covered fraction of each cell, and predicts each region's attributes with a linear head, trained with a binary cross-entropy against class vectors; it uses the same gate and LORO certificate. KP-APN adds to APN-lite, with weight 1, a cross-entropy between the softmax of each prototype map over the $14\times14$ cells and a normalized sum of Gaussians ($\sigma=1$ cell) at the ground-truth visible keypoints of the attribute's region; keypoints are used only in training.

\subsection{Hardware and software}\label{sec:s-hw}
Experiments ran on two machines with the same code and data, checked by SHA-256 checksums: an Apple M4 Pro (PyTorch 2.13.0, MPS backend) and an NVIDIA GB10 (PyTorch 2.14.1, CUDA 13.0). The ResNet-18 FunnyBirds experiments and CUB runs 1--3 were trained on the M4 Pro. CUB run 4 and its rerun, all ResNet-50 and ViT-B/16 runs, CEM, APN-lite under the original APN protocol, the MTSD experiments, and the post hoc retrainings on FunnyBirds (the four full-image baselines with visible-part or all-part targets, the 30-epoch models, the native-scale windows, and the runs that stored per-image outputs) and on CUB (APN-lite on six splits and the additional seeds of the isolated readers and APN-lite) were trained on the GB10. In the matched-keypoint comparison, split~0 was trained on the GB10 and splits 1--5 were trained on the M4 Pro, with all models of a split on the same machine. The retraining of run 4 that saved its models, the matched controls, and the CLIP runs used the GB10; the SUB evaluation, the head sub-regions, the FunnyBirds shared reader, and the tail probe ran on the M4 Pro. A replicate of run 4 on split~0 on the M4 Pro gave 64.1\% ZSL and 15.0\% LORO risk, against 64.5\% and 13.0\% on the GB10. Seeds are fixed, but deterministic kernels were not enabled, and repeated trainings differ (Sec.~\ref{sec:s-cub}).

\section{Additional FunnyBirds results}\label{sec:s-fb}
\begin{table*}[t]\centering\small
\caption{FunnyBirds, unseen classes, with larger recognition backbones (localizer: ResNet-18 in all rows). Certified risk with coverage in parentheses; UCR: unwarranted-commitment rate under decision-critical occlusion; $\lambda$: shortcut index (Sec.~\ref{sec:framework}). Shaded cells mark the lowest risk and, where it is \ocve{}'s, the lowest UCR in each row. CLIP: image encoder of OpenAI CLIP ViT-B/16 with trained heads. $^\ast$With the frozen encoder the joint CBM reaches 0.31 accuracy on seen classes (0.73 fine-tuned), so these rows are not interpreted.}
\label{tab:s-backbones}
\setlength{\tabcolsep}{4pt}
\resizebox{\textwidth}{!}{%
\begin{tabular}{llccccccc}
\toprule
Backbone & Split & \ocve{} & B2-maskpool & Joint CBM & UCR \ocve{} / APN-lite & ZSL \ocve{} / B2 / APN-lite & $\lambda$ \ocve{} / B2\\
\midrule
\multirow{4}{*}{ResNet-50} & 0 & \best{0.65\%} (51.5\%) & 2.78\% (44.5\%) & 87.0\% (50.9\%) & \best{4.6\%} / 34.6\% & 0.797 / 0.681 / 0.774 & $-0.007$ / 0.053\\
& 1 & \best{0.72\%} (52.9\%) & 5.81\% (41.0\%) & 93.6\% (46.3\%) & \best{3.3\%} / 35.1\% & 0.837 / 0.735 / 0.823 & $-0.002$ / 0.052\\
& 2 & \best{0.18\%} (53.7\%) & 5.12\% (41.9\%) & 87.0\% (49.8\%) & \best{4.2\%} / 42.1\% & 0.802 / 0.660 / 0.828 & $-0.002$ / 0.046\\
& mean & \best{0.51\%} (52.7\%) & 4.57\% (42.5\%) & 89.2\% (49.0\%) & \best{4.0\%} / 37.2\% & 0.812 / 0.692 / 0.808 & $-0.004$ / 0.050\\
\midrule
\multirow{4}{*}{ViT-B/16} & 0 & \best{0.37\%} (51.7\%) & 5.19\% (43.1\%) & 94.6\% (53.3\%) & \best{1.1\%} / 18.4\% & 0.799 / 0.710 / 0.785 & $-0.001$ / 0.047\\
& 1 & \best{0.54\%} (52.4\%) & 6.43\% (40.8\%) & 99.2\% (56.2\%) & \best{2.4\%} / 15.3\% & 0.836 / 0.759 / 0.832 & $-0.004$ / 0.056\\
& 2 & \best{0.80\%} (53.3\%) & 2.88\% (46.3\%) & 92.6\% (50.3\%) & \best{3.6\%} / 17.8\% & 0.804 / 0.709 / 0.805 & $-0.002$ / 0.040\\
& mean & \best{0.57\%} (52.5\%) & 4.83\% (43.4\%) & 95.5\% (53.3\%) & \best{2.4\%} / 17.2\% & 0.813 / 0.726 / 0.807 & $-0.002$ / 0.048\\
\midrule
\multirow{4}{*}{\shortstack{CLIP ViT-B/16\\fine-tuned}} & 0 & \best{0.28\%} (50.4\%) & 9.69\% (39.3\%) & 96.4\% (57.8\%) & \best{1.9\%} / 35.2\% & 0.798 / 0.686 / 0.820 & $-0.004$ / 0.040\\
& 1 & \best{0.53\%} (53.4\%) & 4.01\% (42.8\%) & 96.7\% (57.4\%) & \best{2.0\%} / 38.0\% & 0.836 / 0.773 / 0.844 & $-0.002$ / 0.049\\
& 2 & \best{0.44\%} (53.5\%) & 6.05\% (42.5\%) & 91.2\% (51.4\%) & \best{2.9\%} / 24.7\% & 0.803 / 0.721 / 0.828 & $-0.002$ / 0.035\\
& mean & \best{0.42\%} (52.4\%) & 6.58\% (41.5\%) & 94.7\% (55.5\%) & \best{2.2\%} / 32.6\% & 0.812 / 0.727 / 0.830 & $-0.002$ / 0.042\\
\midrule
\multirow{4}{*}{\shortstack{CLIP ViT-B/16\\frozen$^\ast$}} & 0 & \best{5.22\%} (45.6\%) & 11.01\% (38.0\%) & 95.6\% (21.6\%) & 3.7\% / 0.5\% & 0.789 / 0.503 / 0.528 & $-0.001$ / 0.027\\
& 1 & \best{3.09\%} (44.6\%) & 9.49\% (36.1\%) & 96.1\% (21.0\%) & 4.7\% / 0.0\% & 0.825 / 0.457 / 0.531 & $-0.002$ / 0.029\\
& 2 & \best{4.28\%} (45.6\%) & 10.76\% (37.6\%) & 92.9\% (16.9\%) & 6.2\% / 0.9\% & 0.787 / 0.429 / 0.474 & $-0.001$ / 0.014\\
& mean & \best{4.20\%} (45.3\%) & 10.42\% (37.3\%) & 94.9\% (19.8\%) & 4.9\% / 0.5\% & 0.800 / 0.463 / 0.511 & $-0.002$ / 0.023\\
\bottomrule
\end{tabular}}
\end{table*}
\begin{table}[!tb]\centering\small
\caption{FunnyBirds certificates by evidence distance, seen and unseen classes pooled over 3 splits (second training run). Each cell: number of certificates / number wrong.}
\label{tab:s-dx}
\setlength{\tabcolsep}{5pt}
\resizebox{\columnwidth}{!}{%
\begin{tabular}{lcc@{\hskip 14pt}lcc}
\toprule
Visible & $d(x)=1$ & $d(x)\ge2$ & Claimed & $d(x)=1$ & $d(x)\ge2$\\
\midrule
1 & 3 / 2 & 0 / 0 & 2 & 158 / 4 & 0 / 0\\
2 & 82 / 4 & 0 / 0 & 3 & 1{,}131 / 12 & 9 / 0\\
3 & 380 / 3 & 9 / 0 & 4 & 1{,}853 / 5 & 400 / 0\\
4 & 1{,}267 / 4 & 183 / 0 & 5 & 387 / 1 & 786 / 0\\
5 & 1{,}797 / 9 & 1{,}003 / 0 & & &\\
\midrule
All & 3{,}529 / 22 & 1{,}195 / 0 & All & 3{,}529 / 22 & 1{,}195 / 0\\
\bottomrule
\end{tabular}}
\end{table}
\begin{table*}[t]\centering\small
\caption{Agreement decoder versus set decoding on \ocve{}'s claims (FunnyBirds, unseen classes, second training run). Matched: agreement certificates with the most agreeing parts, kept up to \ocve{}'s coverage (ties shared in expectation). Right block: number of agreement certificates and their risk by the number $m$ of agreeing parts.}
\label{tab:s-br1}
\setlength{\tabcolsep}{6pt}
\begin{tabular}{lccccccc}
\toprule
& \ocve{} & Agreement & Matched & $m=2$ & $m=3$ & $m=4$ & $m=5$\\
Split & risk (cov.) & risk (cov.) & risk & & & &\\
\midrule
0 & 0.47\% (51.0\%) & 15.3\% (78.8\%) & 3.3\% & 285: 54.7\% & 424: 20.8\% & 673: 1.5\% & 273: 0.0\%\\
1 & 0.46\% (51.9\%) & 13.8\% (80.2\%) & 2.3\% & 296: 58.4\% & 488: 11.9\% & 591: 0.3\% & 309: 0.0\%\\
2 & 0.63\% (53.0\%) & 17.2\% (78.0\%) & 5.4\% & 309: 57.9\% & 499: 19.6\% & 599: 0.5\% & 230: 0.4\%\\
\midrule
mean & 0.52\% (51.9\%) & 15.4\% (79.0\%) & 3.7\% & & & &\\
\bottomrule
\end{tabular}
\end{table*}
\begin{table}[!tb]\centering\small
\caption{Set-valued certificates on unseen FunnyBirds classes (mean over 3 splits, second training run). Risk: the true class is not in the returned set. APN-lite: top-$s$ sets of its most confident images at \ocve{}'s coverage.}
\label{tab:s-sets}
\resizebox{\columnwidth}{!}{%
\begin{tabular}{lccc}
\toprule
Set size $|D(O)|\le s$ & Coverage & \ocve{} risk$\downarrow$ & APN-lite risk$\downarrow$\\
\midrule
$s=1$ & 51.9\% & \best{0.52\%} & 31.5\%\\
$s=2$ & 66.4\% & \best{0.53\%} & 10.0\%\\
$s=3$ & 70.3\% & \best{0.61\%} & 6.0\%\\
$s=5$ & 77.3\% & \best{0.74\%} & 3.4\%\\
\bottomrule
\end{tabular}}
\end{table}

\paragraph{Backbones.} Table~\ref{tab:s-backbones} gives the per-split results behind the backbone paragraph of the main text (Sec.~\ref{sec:ablate}). Only the recognition models change; the localizer is the ResNet-18 segmenter in every row. With both backbones, the risk order \ocve{} $<$ B2-maskpool $<$ joint CBM holds in each split, but the ratio between the risks of B2-maskpool and \ocve{} varies from 4.3 to 28 with ResNet-50 and from 3.6 to 14 with ViT-B/16. The risk of B2-maskpool falls from 13.8\% with ResNet-18 (Table~\ref{tab:unseen}) to 4.6\% with ResNet-50 and 4.8\% with ViT-B/16, whereas \ocve{} stays between 0.2\% and 0.8\% per split. Among the remaining part errors of B2-maskpool, 70\% (ResNet-50) and 69\% (ViT-B/16) copy a co-occurring value (means over splits), a rate whose chance level is high (see Shortcut rate and chance). Zero-shot accuracy remains similar for \ocve{} and APN-lite (0.81 for both with either backbone). With the CLIP encoder fine-tuned end to end, the order also holds in each split (\ocve{} 0.28--0.53\%, B2-maskpool 4.0--9.7\%, joint CBM 91--97\%), \ocve{} commits on 2.2\% of decision-critical occlusions (APN-lite: 32.6\%), and 68\% of B2-maskpool's part errors copy a co-occurring value. With the encoder frozen and only the heads, prototypes, and global branch trained, as in label-free or post-hoc CBMs, the joint CBM reaches 0.31 accuracy on seen classes (0.73 with the fine-tuned encoder), so its heads learned too little to compare, and we do not interpret this variant; \ocve{}'s risk is 4.2\% there, and APN-lite rarely commits because it also reaches only 0.31 on seen classes.

\paragraph{Evidence distance by stratum.} The main text compares certificates with $d(x)=1$ and $d(x)\ge2$ within strata (Sec.~\ref{sec:stress}). Table~\ref{tab:s-dx} gives the counts. Without strata, the one-sided Fisher test gives $p=0.0016$ ($p=0.006$ on unseen classes alone). Stratifying by the number of truly visible parts, all 22 wrong certificates fall at $d(x)=1$, and the one-sided exact test conditional on the strata gives $p=0.010$. Stratifying by the number of parts the readers claim gives $p=0.11$ ($p=0.11$ on unseen classes alone). The second stratification is the more conservative one, because the number of claims is a model output that also bounds $d(x)$: with two or three claims, $d(x)\ge2$ occurs in only 9 certificates, while 16 of the 22 errors occur there. All three tests are post hoc. Only 36\% of images show two or more distinguishing parts. (The class-level code distance is tested in the seen-class study of Sec.~\ref{sec:s-seen}.)

\paragraph{Agreement decoder.} Table~\ref{tab:s-br1} breaks down the agreement rule of the main text (Sec.~\ref{sec:ablate}, \emph{Decoder and ontology}), which certifies the soft decision when at least two claimed parts match its codeword. Its risk is concentrated in certificates supported by two or three agreeing parts (12--58\%) and falls to 0--1.5\% with four or five. Matching \ocve{}'s coverage therefore keeps mostly the latter, yet the rule remains 5--9 times riskier than set decoding in each split, on the same part claims.

\paragraph{Set-valued certificates.} Table~\ref{tab:s-sets} allows \ocve{} to return compatible sets instead of single classes. With sets of up to three classes, coverage rises from 51.9\% to 70.3\% at 0.6\% set risk (true class outside the set), and APN-lite's top-3 sets miss the true class 6.0\% of the time at that coverage; with up to five classes, coverage reaches 77.3\% while the set risk stays below 1\%. For APN-lite we rank images by confidence and take its top-$s$ classes at the same coverage, where $s$ is the set-size limit.

\begin{table}[!tb]\centering\small
\caption{FunnyBirds, observed risk by evidence distance against an independence-model reference (second training run, seen and unseen classes pooled, per split). $\varepsilon$: average misread rate of \ocve{}'s claims; reference: mean of $\binom{m}{d}\varepsilon^{d}$ over certificates with $m$ claims, a plug-in value of the joint probability of Sec.~\ref{sec:framework} rather than a bound on the risk among certified images.}
\label{tab:s-bound}
\setlength{\tabcolsep}{4pt}
\resizebox{\columnwidth}{!}{%
\begin{tabular}{lccc}
\toprule
& Split 0 & Split 1 & Split 2\\
\midrule
$\varepsilon$ & 0.57\% & 0.68\% & 0.81\%\\
$d(x)=1$: observed / reference & 0.50\% / 2.14\% & 0.60\% / 2.52\% & 0.78\% / 2.96\%\\
$d(x)\ge2$: wrong / expected & 0 / 0.09 & 0 / 0.16 & 0 / 0.19\\
\bottomrule
\end{tabular}}
\end{table}

\paragraph{Independence-model reference (post hoc).} Table~\ref{tab:s-bound} plugs the average misread rate into the bound of Sec.~\ref{sec:framework}, assuming independent errors. The result is an illustrative reference and does not bound certified risk: an average rate does not bound the error rate of every group, and confidence thresholds and decoding select which claims and images are evaluated, which can change how errors depend on each other. The reference exceeds observed risk about fourfold at $d(x)=1$; at $d(x)\ge2$ it corresponds to 0.4 wrong certificates among 1{,}195, and none occurred, so these data cannot test it there.

\paragraph{Shortcut rate and chance (post hoc).} The shortcut rate of the main text counts a wrong part value as copied if it equals the value that co-occurs most, among seen classes, with any other part of the class. Its chance level is high, because a part with few variants has few wrong values left. On the second training run, 75.0\% of the joint CBM's part errors are copies, against 53.2\% expected if the wrong value were drawn uniformly and 59.8\% if it followed variant frequencies (71.5\% vs.\ 48.4\% and 54.6\% when only visible parts count). \ocve{}'s 37--59 part errors per split are copies in 64.0\% of cases, where chance gives 54.3\% and 61.5\%. The shortcut rates of B2-maskpool reported with the backbones (68--70\%) cannot be read without their own chance level, which we did not compute; $\lambda$, which removes the co-occurring part, is the measure we rely on.

\paragraph{$\lambda$ per split and its scale (post hoc).} With ResNet-18 (second training run), $\lambda$ is 0.065, 0.056, and 0.045 for the joint CBM, 0.036, 0.046, and 0.026 for APN-lite, and $-0.005$, $-0.005$, and $-0.002$ for \ocve{} in splits 0--2. As a scale, on the same eligible image--pair cases with part $i$ visible, the joint CBM reads part $j$ as the variant that co-occurs with $i$ in 15.7, 13.0, and 12.2\% of cases, APN-lite does so in 10.6, 11.2, and 8.5\%, and \ocve{} does so in 0.1, 0.2, and 0.3\%; occluding $i$ thus removes about two fifths of the joint CBM's copying.

\begin{table}[!tb]\centering\small
\caption{FunnyBirds test images with and without parts removed by the dataset (second training run, mean over 3 splits). Share: fraction of the test set. Unique: images whose visible parts identify a unique class under perfect reading. Guess: expected accuracy of a perfect reader of the visible parts that guesses uniformly among the classes they allow. Acc.: soft-decoder accuracy over all 50 classes. Risk: certified risk with coverage in parentheses.}
\label{tab:s-removed}
\setlength{\tabcolsep}{2.5pt}
\resizebox{\columnwidth}{!}{%
\begin{tabular}{lccccccc}
\toprule
& & & & \multicolumn{2}{c}{\ocve{}} & \multicolumn{2}{c}{Joint CBM}\\
Test images & Share & Unique & Guess & Acc.$\uparrow$ & Risk$\downarrow$ (cov.) & Acc.$\uparrow$ & Risk$\downarrow$ (cov.)\\
\midrule
Unseen, complete & 59.7\% & 94.1\% & 0.970 & 0.917 & 0.39\% (73.8\%) & 0.133 & 90.0\% (51.0\%)\\
\quad official test set & 4.8\% & 93.7\% & 0.968 & 0.910 & 0.48\% (72.3\%) & 0.153 & 89.3\% (46.7\%)\\
Unseen, parts removed & 40.3\% & 28.2\% & 0.413 & 0.320 & 1.21\% (19.5\%) & 0.027 & 97.6\% (41.5\%)\\
\midrule
Seen, complete & 57.5\% & 91.2\% & 0.955 & 0.823 & 0.16\% (70.4\%) & 0.933 & 6.0\% (97.2\%)\\
Seen, parts removed & 42.5\% & 28.3\% & 0.410 & 0.326 & 1.22\% (18.5\%) & 0.374 & 45.3\% (53.6\%)\\
\bottomrule
\end{tabular}}
\end{table}

\paragraph{Removed parts and complete birds (post hoc).} FunnyBirds removes one to five parts from 41.3\% of its 50{,}000 training images (each part from about 24.6\% of all training images) and trains its reference models with multi-label targets; its 500 official test images are complete~\cite{hesse2023funnybirds}. Our unseen-class test sets contain 100 official test images and 2{,}000 training-split images, and our seen-class test sets contain only training-split images, so about 40\% of them lack at least one part. Table~\ref{tab:s-removed} splits the outputs of the second training run accordingly. On complete birds, perfect readers of the visible parts identify a unique class in 94\% of unseen-class and 91\% of seen-class images, and by guessing among the classes those parts allow they would be right on 97\% and 96\% of them (the perfect readers of Sec.~\ref{sec:mechanism}); the joint CBM reaches 0.933 seen-class accuracy and reads visible seen-class parts at 0.970--0.996, yet its unseen-class certificates are wrong 90\% of the time, and its accuracy on visible unseen-class parts ranges from 0.24 (tail) to 0.93 (foot). On images with removed parts, perfect readers identify a unique class in only 28\% of cases, which caps every certificate. \ocve{}'s seen-class accuracy trails the joint CBM's mostly on complete birds (0.823 vs.\ 0.933), which account for three quarters of the gap over all seen-class images; since perfect readers would reach 0.955 there, the gap comes mainly from the localizer and readers rather than from reporting hidden parts as $\bot$. The seen-class corruption study (Table~\ref{tab:s-seen-corr}) has the same structure: 34\% of its 2{,}500 test images lack parts; with the visibility flags of that study, perfect readers identify a unique class in 97.0\% of the complete birds and 30.1\% of the rest (94.2\% and 28.0\% when computed from ground-truth part maps, as in Table~\ref{tab:s-removed}), and the black box of its first two columns reaches 0.933 on complete birds and 0.936 on the 500 official test images. Hesse et al. report 0.99--1.00 for VGG16 and ResNet-50 trained on all 50{,}000 images (ResNet-50 for 120 epochs)~\cite{hesse2023funnybirds}; our models are ResNet-18s trained for 3--5 epochs on 12{,}000 images with single-label targets.

\begin{table}[!tb]\centering\small
\caption{Joint CBM retrained with class-level targets for every part, as in the main text, or with the concept loss restricted to parts visible in the image (unseen FunnyBirds classes, ResNet-18, mean over 3 splits). Gate: parts that the localizer of the second training run does not find are reported as $\bot$. Risk: certified risk with coverage in parentheses. Seen: seen-class accuracy on complete birds, without gate. $\lambda$ uses argmax readings and does not depend on the gate.}
\label{tab:s-b2vis}
\setlength{\tabcolsep}{2.5pt}
\resizebox{\columnwidth}{!}{%
\begin{tabular}{lccccccc}
\toprule
& & \multicolumn{2}{c}{No gate} & \multicolumn{2}{c}{Gate} & &\\
Targets & Epochs & Risk$\downarrow$ (cov.) & GZSL$\uparrow$ & Risk$\downarrow$ (cov.) & GZSL$\uparrow$ & Seen$\uparrow$ & $\lambda$ (splits 0--2)\\
\midrule
All parts & 5 & 88.8\% (50.1\%) & 0.098 & 82.4\% (42.0\%) & 0.174 & 0.934 & 0.054 / 0.053 / 0.038\\
All parts & 30 & 94.4\% (52.3\%) & 0.071 & 91.3\% (37.7\%) & 0.145 & 0.947 & 0.062 / 0.065 / 0.054\\
Visible parts & 5 & 72.1\% (41.6\%) & 0.211 & 60.1\% (36.2\%) & 0.302 & 0.936 & 0.040 / 0.045 / 0.034\\
Visible parts & 30 & 73.1\% (42.2\%) & 0.215 & 63.0\% (35.4\%) & 0.314 & 0.947 & 0.047 / 0.048 / 0.036\\
\bottomrule
\end{tabular}}
\end{table}

\paragraph{Joint CBM trained longer or on visible parts (post hoc).} Like B2-ind, CEM, and APN-lite, the joint CBM learns class-level values for parts that the image does not show (Sec.~\ref{sec:protocol}). We therefore retrained it with the main recipe and a new seed, for 5 or 30 epochs, either with class-level targets for every part or with each head's loss restricted to images in which its part covers at least 15 pixels of the ground-truth part map. The threshold $\tau$ is the 1\% quantile of the top probability of correct claims on visible parts of seen-class validation images, shared by the four models, and each model is scored with and without the localizer gate (Table~\ref{tab:s-b2vis}). We count the failure as persisting if certified risk, taking the lower-risk scoring of each model, stays at or above 20\% after 30 epochs and with the visible-part loss, and if $\lambda$ of the 5-epoch visible-part model stays at or above 0.02 in every split; the test counts only if both 30-epoch models reach 0.90 seen-class accuracy on complete birds. All conditions hold: the risks are 91.3\% and 60.1\%, $\lambda$ is 0.034--0.045, and both 30-epoch models reach 0.947. At 5 epochs, the visible-part loss lowers risk by 17 points without the gate and by 22 points with it. The visible-part models still read visible tails of unseen classes correctly in only 42\% (5 epochs) and 50\% (30 epochs) of cases, against 26\% for the class-level 5-epoch model, whereas \ocve{}'s readers are right on 97--99\% of the tail claims they make.

\paragraph{Training length and compute (post hoc).} Seen-class accuracy in Table~\ref{tab:unseen} counts all seen-class test images, 42.5\% of which lack at least one part; guessing among the classes that the visible parts allow, the perfect readers of Table~\ref{tab:s-removed} would reach 0.72 over these images (0.955 on complete birds and 0.41 on the rest). To test whether the failure of the concept models reflects a short training schedule, we also trained CEM for 30 epochs on the three splits, changing only the number of epochs. On complete seen-class birds, its soft-decoded accuracy rises from 0.938 (5 epochs, retrained to store per-image outputs) to 0.942. Its certified risk on unseen classes falls from 93.8\% to 91.1\% (89.7, 93.8, and 89.8\% per split), while its coverage rises from 45.5\% to 58.9\%, so it issues more wrong certificates, and $\lambda$ rises from 0.051 to 0.056. The 30-epoch joint CBM above reaches 0.947 on complete birds, with 94.4\% risk and $\lambda$ of 0.054--0.065. Both 30-epoch models are within 0.02 of the perfect-reader accuracy on complete birds (0.955), and their certified risk stays above 90\%. Both remain below the 0.99--1.00 that Hesse et al.~\cite{hesse2023funnybirds} report for VGG16 and ResNet-50 trained on about four times as many images. Counted with the FLOP counter of PyTorch, one forward pass costs 3.6 GFLOPs for the joint CBM and CEM at $224^2$, 5.5 GFLOPs for the localizer at $256^2$, and 0.67 GFLOPs for a reader at $96^2$. With backward passes counted as twice the forward cost and 3.5 reader crops per training image, training \ocve{} costs 1.4 PFLOPs (localizer 1.0, readers 0.4), 2.2 times the 0.65 PFLOPs of the 5-epoch joint CBM, while the 30-epoch joint CBM costs 3.9 PFLOPs, 2.8 times as much as \ocve{}; training B2-ind costs 3.3 PFLOPs. At inference, \ocve{} needs 7.8 GFLOPs per image on average and 8.8 with all five readers. This is 2.2 and 2.4 times the cost of the joint CBM, which equals one ResNet-18 pass (3.6 GFLOPs; Sec.~\ref{sec:method}), and 0.43 and 0.49 times that of B2-ind (18.1 GFLOPs).

\begin{table}[!tb]\centering\small
\caption{B2-ind (one ResNet-18 per part on the full image) retrained with class-level targets for every part or only on images that show the part, scored without and with the localizer gate (unseen FunnyBirds classes, mean over 3 splits; post hoc). Risk: certified risk with coverage in parentheses. UCR: commitment rate under decision-critical occlusion, by certificate (with gate) and by confidence. $\lambda$ uses argmax readings. Shading: lower risk and higher accuracy.}
\label{tab:s-b2ind}
\setlength{\tabcolsep}{2.5pt}
\resizebox{\columnwidth}{!}{%
\begin{tabular}{lccccc}
\toprule
Targets & Risk$\downarrow$, no gate & Risk$\downarrow$, gate & GZSL / ZSL$\uparrow$ (gate) & $\lambda$ (splits 0--2) & UCR$\downarrow$ cert.\ / conf.\\
\midrule
All parts & 27.6\% (40.9\%) & 13.0\% (44.9\%) & 0.619 / 0.831 & 0.011 / 0.019 / 0.007 & 33.3\% / 28.9\%\\
Visible parts & \best{16.3\% (42.0\%)} & \best{5.6\% (49.3\%)} & \best{0.698 / 0.841} & 0.007 / 0.008 / 0.004 & 30.5\% / 2.0\%\\
\bottomrule
\end{tabular}}
\end{table}

\paragraph{Full-image part readers trained on visible parts (post hoc).} Table~\ref{tab:s-b2ind} repeats the analysis above for B2-ind, with the splits, data, and segmenter of the second training run and 5 epochs. The gate lowers certified risk from 27.6\% to 13.0\%, and the visible-part loss lowers it further to 5.6\% (6.4, 5.0, and 5.5\% per split), which is 11 times \ocve{}'s risk in the same run (0.52\%) at similar coverage; the accuracy is slightly higher than \ocve{}'s (GZSL 0.698 vs.\ 0.677). The seen-class accuracy of the visible-part model on complete birds is 0.93--0.95. Its $\lambda$ stays at 0.004--0.008, close to \ocve{}'s and far below the joint CBM's (0.045--0.065). Of the 20 ordered pairs that $\lambda$ averages, only $\lambda_{\mathrm{tail}\leftarrow\mathrm{wing}}$ exceeds 0.026 (0.097, 0.080, and 0.059 per split). Decoded among the unseen classes only, the gated models certify at 1.02\% (all parts) and 1.24\% (visible parts) risk at 50.1\% and 56.6\% coverage, 2.8 and 3.4 times \ocve{}'s 0.36\% at 71.1\%. Under decision-critical occlusion, the certificates of the visible-part model commit on 30.5\% of images, against 2.3\% for \ocve{}, while its confidence commits on only 2.0\%. Its confidence threshold, set for 1\% risk on seen validation images, lies at 0.9999--1.0 and also admits only 8.6--32.3\% of clean unseen-class images (19\% on average), at 0--6.2\% risk. Its readers still read the visible tails of unseen classes correctly in only 73\% of cases, whereas \ocve{}'s readers are right on 97--99\% of the tail claims they make.

\paragraph{CEM and APN-lite trained on visible parts (post hoc).} To give every full-image baseline of Table~\ref{tab:unseen} the supervision of \ocve{}, we retrained CEM and APN-lite like B2-ind above, with the splits, data, and segmenter of the second training run: CEM's concept loss and interventions cover only the parts visible in each image, and APN-lite's prototype regression covers only the attribute slots of visible parts. Claims use $\tau_g$ set on visible parts of the validation images (for APN-lite, on a softmax over the prototype scores of each group), and parts that the localizer misses are $\bot$. With the gate, CEM certifies at 71.7\% risk (64.6, 80.9, and 69.5\% per split) at 36.7\% coverage, and APN-lite certifies at 39.7\% (39.0, 50.3, and 29.7\%) at 34.5\%, against 0.52\% for \ocve{}; without the gate, their risks are 80.8\% and 57.5\%. Reruns with targets for every part certify at 86.3\% and 58.0\% with the gate. Without the gate, CEM certifies at 92.5\%, and APN-lite, scored without $\tau_g$ as in Table~\ref{tab:unseen}, certifies at 75.0\%. Training on visible parts leaves $\lambda$ at 0.047 for CEM and 0.028 for APN-lite (\ocve{}: $-0.004$), and both models reach 0.94 accuracy on complete birds of seen classes. APN-lite's most confident predictions at \ocve{}'s coverage are wrong 27.5\% of the time (26.0\% for the rerun).

\paragraph{Tail-occlusion probe (post hoc).} To test whether the $\times3$ wing reader of A1 uses the tail it sees, we retrained it on the localizer of the second training run with the same seed. The probe uses unseen-class test images with a detected wing and a visible tail whose true wing differs from the wing variant that co-occurs most with the true tail (785, 821, and 719 images per split). For each reader, $\lambda_{\mathrm{wing}\leftarrow\mathrm{tail}}$ is the share of wing readings equal to that variant on the original crop minus the same share when the tail is grayed (ground-truth mask) before cropping. The tail lies in 99.0--99.5\% of these $\times3$ wing crops. $\lambda_{\mathrm{wing}\leftarrow\mathrm{tail}}$ is 0.11 points on average (0.25, 0.37, and $-0.28$ per split) for the $\times3$ reader, 0.0 points for \ocve{}'s masked wing reader, and 4.0 points (5.2, 3.5, and 3.3) for the joint CBM.

\paragraph{Where B2-ind looks when it misreads (post hoc).} We computed Grad-CAM maps (layer3, $14\times14$, for the predicted variant) of the visible-part B2-ind readers on the unseen-class test images, for every part that is visible in the ground-truth part map (22{,}023 readings over three splits). For correct readings, the peak of the map lies on the part, whose mask we dilate by 16\,px, in 85.6\% of cases; for wrong readings, it does so in 30.0\% (27.3--31.7\% per split). The peaks of wrong readings fall on another of the five parts in 14\% of cases and on the body or the background in 60\% (the dilated regions overlap). Wrong readings concentrate on small visible areas (median 38 vs.\ 171\,px at $224^2$), and the tail accounts for 66\% of them, but the gap persists for parts of at least 150\,px (40.0\% vs.\ 88.5\%). The analysis is correlational, since a diffuse map may also follow from an uncertain reading.

\paragraph{Native-scale windows (post hoc).} To separate the local window from the magnification of the part, each reader receives the full image resized to $224^2$, with every pixel outside the part's $\times1$ or $\times3$ square box of A1 grayed, so the part keeps its native scale and position. Readers, labels, $\tau$, and decoding are as in A1. The $\times1$ window certifies at 0.91\% risk (1.35, 0.38, and 0.99\% per split) at 51.0\% coverage, and the $\times3$ window certifies at 1.10\% at 50.4\% coverage, against 1.0\% for the magnified $\times1$ crop. $\lambda$ is $-0.003$ to 0.001 for both, and they commit on 6.4\% and 15.0\% of decision-critical occlusions (\ocve{}: 2.3\%).

\paragraph{Full-image readers given a window at test (post hoc).} To test whether the visible-part B2-ind (Table~\ref{tab:s-b2ind}) errs because of the context it sees at test time, its trained readers received the same native-scale windows, with $\tau_g$ set again on validation windows and $\bot$ for parts without a predicted mask. On the full image with the gate, they reproduce Table~\ref{tab:s-b2ind} (5.6\% risk at 49.3\% coverage). With $\times3$ windows, their accuracy on visible parts stays at 0.93, and certified risk is 4.5\% (4.5, 3.8, and 5.1\% per split) at 49.0\% coverage. With $\times1$ windows, accuracy on visible parts falls to 0.86 and risk rises to 7.2\% at 36.6\% coverage; these inputs differ too much from the training images to interpret.

\paragraph{Ground-truth masks at test (post hoc).} With the readers and $\tau$ of A3, crops from ground-truth masks give 146 wrong certificates of 2{,}923 (risk 2.9, 4.9, and 6.9\% per split), against 17 of 3{,}272 with the localizer's masks. All 146 involve misreads of parts that both masks find; parts that only the ground-truth masks find yield 3 claims in total. On parts found by both, the readers misread 4.0\% of the ground-truth-mask crops and 0.65\% of the crops from predicted masks, on which they were trained, and the misread parts in wrong certificates are small (median ground-truth area 75 pixels, against 251 for all claims).

\paragraph{Calibrated stacking (post hoc).} On pooled seen and unseen test images, at \ocve{}'s coverage there (50.8\%), APN-lite's most confident images are wrong 3.7\% of the time with the offset that maximizes $H$ and 3.2\% with the offset that minimizes its risk, both tuned on the test images; \ocve{}'s certificates are wrong 0.47\% of the time.

\paragraph{Ontology corruption.} Each (class, part) entry of the code is replaced, with probability $p$, by a different variant drawn uniformly; \ocve{}'s stored claims are re-decoded with the set decoder, and risk is averaged over five corruption seeds per split and over the three splits.

\begin{figure*}[t]\centering
\includegraphics[width=\textwidth]{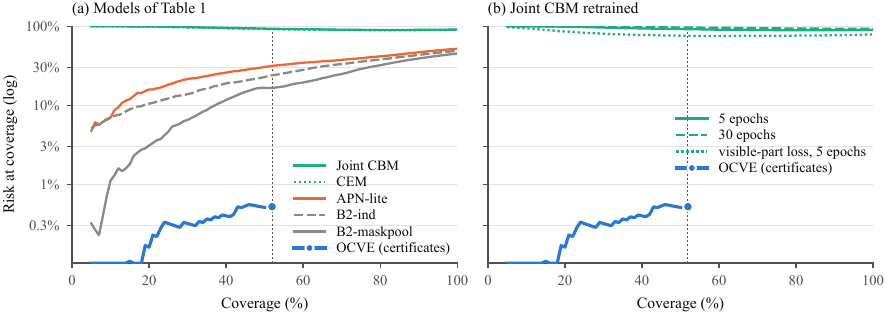}
\caption{Risk--coverage on unseen FunnyBirds classes (mean over 3 splits; post hoc). Each model ranks its predictions over all 50 classes by its own confidence (for the concept models, the class probability propagated from their concept probabilities); \ocve{} ranks its certificates by $d(x)$ and abstains on the remaining images. Dotted line: \ocve{}'s coverage; zero risk is drawn at the lower edge. (a) Models of Table~\ref{tab:unseen}. \ocve{}, the joint CBM, and APN-lite come from the second training run; B2-ind, B2-maskpool, and CEM were retrained with the same protocol and seeds to store per-image outputs, and their certified risks (33.4\%, 13.3\%, and 93.8\%) are close to those in Table~\ref{tab:unseen}. (b) Joint CBM retrained for 5 or 30 epochs or with the concept loss restricted to visible parts (Table~\ref{tab:s-b2vis}), without gate.}
\label{fig:s-riskcov}
\end{figure*}

\paragraph{Risk--coverage of every model (post hoc).} Table~\ref{tab:unseen} compares certificates at the coverage each model reaches. Figure~\ref{fig:s-riskcov} instead ranks the predictions of every model by its own confidence and compares the models at every coverage. At \ocve{}'s coverage (51.9\%), the most confident predictions are wrong 16.6\% of the time for B2-maskpool (15.0--18.9\% per split), 23.9\% for B2-ind (16.1--35.0\%), 31.5\% for APN-lite, 91.7\% for CEM, and 93.2\% for the joint CBM, whereas \ocve{}'s certificates are wrong 0.52\% of the time. At 25\% coverage, the same rankings give 4.6\%, 12.2\%, 18.3\%, 97.8\%, and 98.7\%, against 0.32\% for \ocve{}. The joint CBM is wrong on all of its 10\% most confident predictions. Without gate, the retrained joint CBMs of Table~\ref{tab:s-b2vis} reach 92.2\% (5 epochs), 96.6\% (30 epochs), and 75.8\% (visible-part loss) at \ocve{}'s coverage.

\paragraph{Unseen-only decoding (post hoc).} FunnyBirds certificates in Table~\ref{tab:unseen} are decoded against all 50 classes. Decoding the outputs of the second training run against the unseen classes only, \ocve{} certifies 71.1\% of unseen images at 0.36\% risk, and the joint CBM certifies only 4.8\%, at 25.6\% risk; APN-lite's most confident images at \ocve{}'s coverage are wrong 3.0\% of the time. For the joint CBM retrained with the main recipe (Table~\ref{tab:s-b2vis}, class-level targets, 5 epochs), 99.8\% of the wrong certificates on unseen-class images name a seen class (2{,}798 of 2{,}805). The certificates of B2-maskpool, B2-ind, and CEM, retrained to store per-image outputs (Fig.~\ref{fig:s-riskcov}), reach 1.9\% risk at 55.0\% coverage (1.6--2.5\% per split), 4.8\% at 23.1\%, and 32\% at 4.1\%. A further retraining of B2-maskpool that stores per-class scores (post hoc; certified risk 14.4\% at 36.6\% coverage) reaches 2.4\% with its unseen-only certificates at 55.1\% coverage and 2.7\% when its most confident images are taken at \ocve{}'s unseen-only coverage (\ocve{}: 0.36\%). With calibrated stacking on pooled seen and unseen images, at \ocve{}'s coverage there, it reaches 3.5\% with the offset that maximizes $H$ and 3.0\% with the offset that minimizes its risk (\ocve{}: 0.47\%). Its $\lambda$ is 0.008 (0.004--0.010 per split), and it commits on 16.8\% of decision-critical occlusions.

\paragraph{Mask leakage controls.} Readers retrained on unmasked part bounding boxes retain 99.9\% of \ocve{}'s GZSL advantage over the joint CBM, and a probe that sees only one part's predicted binary mask predicts other parts 0.3 points \emph{below} the co-occurrence prior, so this probe found no exploitable cross-part signal in the masks. The probe is a ResNet-18 with one head per other part, trained for three epochs on the binary mask crop of one part.

\paragraph{Shared reader (post hoc).} One ResNet-18 that reads the masked crops of all five parts in separate forward passes, with one head per part, certifies at 0.74, 0.81, and 1.26\% risk in splits 0--2 (0.94\% at 52.4\% coverage on average), against 0.47, 0.46, and 0.63\% for separate readers on the same localizer, crops, and images, so its risk is 1.6--2.0 times theirs in each split (1.8 times on average); its ZSL accuracy is 0.808 (separate readers: 0.811).

\paragraph{Distance to seen classes.} The joint CBM's certified risk is 79--94\% at distance 1 and 94--97\% at distance $\ge2$, while \ocve{} stays at 0--0.8\% in both.

\section{Additional CUB results}\label{sec:s-cub}
\begin{table*}[t]\centering\small
\caption{Matched keypoint supervision on CUB, per split (class-level attributes, ResNet-18). ZSL: per-class accuracy among the 50 unseen species. ISO: isolated readers. Risk: LORO certificates of ISO; the other models keep their top-ranked images at ISO's coverage (APN-lite and KP-APN by confidence, POOL by its own LORO certificate and then margin). POOL: features of a shared trunk pooled in the boxes that ISO crops. Shaded cells mark the best value per row within each block.}
\label{tab:s-taskA}
\setlength{\tabcolsep}{4pt}
\resizebox{\textwidth}{!}{%
\begin{tabular}{l|cccc|c|cccc|cccc|c|cccc}
\toprule
& \multicolumn{9}{c|}{Clean images} & \multicolumn{9}{c}{Head grayed out}\\
& \multicolumn{4}{c|}{ZSL$\uparrow$} & Cov. & \multicolumn{4}{c|}{Risk$\downarrow$} & \multicolumn{4}{c|}{ZSL$\uparrow$} & Cov. & \multicolumn{4}{c}{Risk$\downarrow$}\\
Split & ISO & POOL & APN & KP-APN & & ISO & POOL & APN & KP-APN & ISO & POOL & APN & KP-APN & & ISO & POOL & APN & KP-APN\\
\midrule
0 & \best{64.9} & 54.1 & 59.4 & 57.2 & 41.9 & \best{13.4} & 27.0 & 22.9 & 21.9 & 32.9 & 30.5 & \best{33.1} & 29.2 & 18.4 & 36.2 & 42.2 & \best{32.1} & 39.2\\
1 & \best{61.4} & 52.0 & 55.2 & 52.9 & 43.7 & \best{16.6} & 27.6 & 24.7 & 25.5 & \best{29.8} & 27.3 & 28.7 & 27.8 & 14.6 & 41.2 & \best{37.0} & 42.7 & 44.5\\
2 & \best{61.8} & 53.7 & 54.4 & 53.1 & 44.3 & \best{22.5} & 32.3 & 27.8 & 27.6 & \best{28.4} & 24.2 & 27.5 & 25.7 & 19.8 & 55.2 & 51.0 & 47.2 & \best{46.2}\\
3 & \best{62.1} & 52.9 & 54.3 & 52.7 & 42.2 & \best{20.6} & 30.6 & 27.7 & 29.5 & 26.6 & 24.6 & \best{27.0} & 25.7 & 18.8 & 50.4 & 60.7 & \best{49.3} & \best{49.3}\\
4 & \best{63.6} & 50.5 & 51.5 & 52.4 & 43.9 & \best{19.6} & 31.9 & 31.8 & 27.7 & \best{33.6} & 28.1 & 29.1 & 25.2 & 21.6 & 46.8 & 50.6 & \best{43.0} & 52.8\\
5 & \best{55.8} & 48.2 & 51.7 & 49.1 & 42.8 & \best{25.8} & 32.6 & 27.0 & 30.5 & 28.2 & \best{28.4} & 28.1 & 27.1 & 17.0 & 43.7 & 49.4 & \best{40.5} & 45.3\\
\midrule
0--2 & \best{62.7} & 53.3 & 56.4 & 54.4 & 43.3 & \best{17.5} & 28.9 & 25.1 & 25.0 & \best{30.4} & 27.3 & 29.8 & 27.6 & 17.6 & 44.2 & 43.4 & \best{40.7} & 43.3\\
3--5 & \best{60.5} & 50.6 & 52.5 & 51.4 & 43.0 & \best{22.0} & 31.7 & 28.8 & 29.2 & \best{29.5} & 27.0 & 28.1 & 26.0 & 19.1 & 47.0 & 53.6 & \best{44.3} & 49.1\\
all & \best{61.6} & 51.9 & 54.4 & 52.9 & 43.2 & \best{19.8} & 30.3 & 27.0 & 27.1 & \best{29.9} & 27.2 & 28.9 & 26.8 & 18.4 & 45.6 & 48.5 & \best{42.5} & 46.2\\
\bottomrule
\end{tabular}}
\end{table*}
\begin{table}[!tb]\centering\small
\caption{CUB run 4 and a rerun with the same code and seeds (ResNet-18, class-level attributes, clean images). ZSL and LORO risk of the isolated readers (ISO) and APN-lite at ISO's coverage.}
\label{tab:s-rerun}
\setlength{\tabcolsep}{3pt}
\resizebox{\columnwidth}{!}{%
\begin{tabular}{l|ccc|ccc}
\toprule
& \multicolumn{3}{c|}{Run 4} & \multicolumn{3}{c}{Rerun}\\
Split & ZSL ISO / APN & Cov. & Risk ISO / APN & ZSL ISO / APN & Cov. & Risk ISO / APN\\
\midrule
0 & 64.5 / 59.2 & 40.5 & 13.0 / 20.3 & 63.7 / 59.0 & 41.2 & 16.3 / 20.5\\
1 & 62.4 / 51.5 & 43.2 & 16.9 / 24.4 & 61.5 / 52.2 & 45.2 & 20.6 / 25.5\\
2 & 62.5 / 55.1 & 44.9 & 23.1 / 27.4 & 62.1 / 54.6 & 44.4 & 23.7 / 27.1\\
\midrule
mean & 63.1 / 55.3 & 42.9 & 17.7 / 24.0 & 62.5 / 55.3 & 43.6 & 20.2 / 24.3\\
PS & 62.2 / 58.1 & 41.5 & 18.2 / 20.5 & 63.1 / 57.5 & 40.1 & 17.3 / 17.8\\
\bottomrule
\end{tabular}}
\end{table}
\begin{table}[!tb]\centering\small
\caption{CUB run 4 with ResNet-50 recognition models (class-level attributes, clean images). Risk: LORO for ISO; APN-lite and the joint reader at ISO's coverage. Shaded cells mark the best value per row and metric.}
\label{tab:s-r50}
\setlength{\tabcolsep}{3pt}
\resizebox{\columnwidth}{!}{%
\begin{tabular}{l|ccc|c|ccc}
\toprule
& \multicolumn{3}{c|}{ZSL$\uparrow$} & & \multicolumn{3}{c}{Risk$\downarrow$}\\
Split & ISO & APN-lite & Joint & Cov. & ISO & APN-lite & Joint\\
\midrule
0 & \best{70.7} & 62.8 & 57.4 & 47.2 & \best{10.3} & 22.4 & 24.1\\
1 & \best{64.4} & 54.9 & 56.8 & 45.3 & \best{18.2} & 30.0 & 22.0\\
2 & \best{65.9} & 55.3 & 55.6 & 49.0 & \best{20.2} & 28.7 & 28.7\\
\midrule
mean & \best{67.0} & 57.6 & 56.6 & 47.2 & \best{16.2} & 27.0 & 24.9\\
PS & \best{66.3} & 58.8 & 57.2 & 43.4 & \best{15.2} & 19.2 & 25.6\\
\bottomrule
\end{tabular}}
\end{table}
\begin{table}[!tb]\centering\small
\caption{Generalized zero-shot learning on CUB (rerun of run 4, ResNet-18, class-level attributes). AUSUC: area under the seen--unseen curve; $H$: harmonic mean of per-class seen and unseen accuracy, at its best $\gamma$ and at $\gamma=0$. Shaded cells mark the best value per row within each block.}
\label{tab:s-ausuc}
\setlength{\tabcolsep}{3pt}
\resizebox{\columnwidth}{!}{%
\begin{tabular}{l|ccc|ccc|ccc}
\toprule
& \multicolumn{3}{c|}{AUSUC$\uparrow$} & \multicolumn{3}{c|}{Best $H\uparrow$} & \multicolumn{3}{c}{$H$ at $\gamma=0\uparrow$}\\
Split & ISO & APN & Joint & ISO & APN & Joint & ISO & APN & Joint\\
\midrule
0 & \best{0.422} & 0.403 & 0.266 & \best{0.567} & 0.538 & 0.432 & 0.386 & \best{0.461} & 0.223\\
1 & \best{0.431} & 0.366 & 0.265 & \best{0.577} & 0.516 & 0.434 & 0.360 & \best{0.437} & 0.210\\
2 & \best{0.427} & 0.373 & 0.249 & \best{0.563} & 0.518 & 0.413 & 0.378 & \best{0.400} & 0.158\\
\midrule
mean & \best{0.427} & 0.381 & 0.260 & \best{0.569} & 0.524 & 0.426 & 0.374 & \best{0.433} & 0.197\\
PS & \best{0.426} & 0.399 & 0.255 & \best{0.576} & 0.534 & 0.424 & 0.381 & \best{0.410} & 0.222\\
\bottomrule
\end{tabular}}
\end{table}

\paragraph{Matched keypoint supervision per split.} Table~\ref{tab:s-taskA} lists the six splits of the matched-keypoint comparison (Sec.~\ref{sec:cub}). Splits 0--2 are the development splits; splits 3--5 were drawn afterwards and run with the code unchanged. On clean images the isolated readers (ISO) lead all three competitors in accuracy and in LORO risk at equal coverage in every split. The smallest risk margin occurs in split~5 (25.8\% vs.\ 27.0\% for APN-lite). With the head grayed out, the four models are within 3.1 points in mean accuracy, and APN-lite's confidence ranks errors best (42.5\% vs.\ 45.6\% for ISO, mean over six splits), as in run 4.

\paragraph{Run-to-run variability.} The isolated readers of run 4 were trained five times on splits 0--2 with the same code and seeds: run 4 itself, the ISO model of the matched-keypoint comparison, a rerun that also stored outputs on seen classes (used for AUSUC below), a retraining that saved its models (used for the matched controls, SUB, and the head sub-regions), and, for split~0 only, a replicate on different hardware (Sec.~\ref{sec:s-hw}). Table~\ref{tab:s-rerun} compares run 4 with the rerun. Across trainings of the same split, accuracy changes by at most 2.5 points, whereas LORO risk changes by up to 4.2 points (split~1: 16.6\% in the matched-keypoint run, 16.9\% in run 4, 20.6\% in the rerun, and 20.8\% in the retraining). Training is not deterministic on our hardware (Sec.~\ref{sec:s-hw}); we did not isolate which source of variation causes these differences. Over all comparisons of ISO with APN-lite on clean images (run 4, the rerun, the retraining, ResNet-50, and the six matched-keypoint splits; 21 split-level comparisons including the proposed split, which share data and are not independent replications), ISO's LORO risk is lower in every case, and its mean over splits is 0.60--0.83 times APN-lite's: 0.74 (run 4), 0.83 (rerun), 0.74 (retraining), 0.70 (matched keypoints, splits 0--2), 0.76 (splits 3--5), and 0.60 (ResNet-50).

\paragraph{ResNet-50.} Table~\ref{tab:s-r50} reports run 4 with ResNet-50 recognition models; the keypoint model stays ResNet-18. The isolated readers lead APN-lite and the joint reader in accuracy in every split and on the proposed split, and their LORO risk is lower than both in every split. The joint reader gains the most from the larger backbone (0.479 to 0.566 ZSL), the isolated readers gain 3.9 points, and APN-lite gains 2.3 points. Of the three results that did not favor isolation with ResNet-18, two persist and one reverses. With the head grayed out, APN-lite is slightly more accurate (0.359 vs.\ 0.342) and ranks errors better (36.7\% vs.\ 39.5\% risk). With image-level labels, the isolated readers now reach 0.623 vs.\ 0.576, with 16.7\% LORO risk vs.\ 24.1\% (APN-lite) and 34.6\% (joint reader). Over all 200 classes at zero calibration offset, the isolated readers still trail APN-lite (0.270 vs.\ 0.304).

\paragraph{AUSUC.} At zero calibration offset, the isolated readers trail APN-lite over all 200 classes (Sec.~\ref{sec:cub}). To separate calibration from discrimination, we computed the area under the seen--unseen accuracy curve (AUSUC)~\cite{chao2016empirical}: a constant $\gamma$ is subtracted from the scores of all seen classes, and the unseen and seen per-class accuracies are traced exactly over all breakpoints of $\gamma$. This requires outputs on seen-class test images, which run 4 did not store, so the numbers come from the rerun above (clean images, class-level attributes). Table~\ref{tab:s-ausuc} shows that the isolated readers have the larger AUSUC and the larger best harmonic mean in every split, while APN-lite has the higher harmonic mean at $\gamma=0$. The deficit at $\gamma=0$ is therefore consistent with an offset between seen and unseen scores (the isolated readers favor seen classes: $s=0.737$, $u=0.251$) rather than with weaker discrimination.

\begin{table}[!tb]\centering\small
\caption{CUB development runs 1--3 on splits 0--2 (mean over splits), which preceded run 4. Risk: certificates of the isolated readers when two or more regions agree, and of the other models at the same coverage (36--43\%). Occl.: ZSL with the head grayed out. $\bot$ in reader: \textsc{absent} output in each reader (run 3). Shading: best value per run; run 1 is a localization diagnostic and is not shaded.}
\label{tab:s-cubruns}
\setlength{\tabcolsep}{2.5pt}
\resizebox{\columnwidth}{!}{%
\begin{tabular}{cllccc}
\toprule
Run & Model & Sup. & ZSL$\uparrow$ & Risk$\downarrow$ & Occl.$\uparrow$\\
\midrule
\multirow{4}{*}{1} & Joint attribute reader & img & 0.396 & -- & --\\
& Isolated, keypoints v1 & img & 0.400 & -- & --\\
& Isolated, GT keypoints & img & 0.574 & -- & --\\
& APN-lite & cls & 0.545 & -- & --\\
\midrule
\multirow{3}{*}{2} & Joint attribute reader & img & 0.399 & 45.1\% & 0.228\\
& Isolated, keypoints v2 & img & \best{0.566} & 28.2\% & 0.231\\
& APN-lite & cls & 0.552 & \best{23.1\%} & \best{0.290}\\
\midrule
\multirow{5}{*}{3} & Joint attribute reader & img & 0.403 & 45.3\% & 0.224\\
& Joint attribute reader & cls & 0.482 & 33.1\% & 0.284\\
& Isolated + $\bot$ in reader & img & 0.517 & 35.6\% & 0.248\\
& Isolated + $\bot$ in reader & cls & 0.510 & 36.9\% & 0.240\\
& APN-lite & cls & \best{0.552} & \best{23.5\%} & \best{0.290}\\
\bottomrule
\end{tabular}}
\end{table}

\begin{figure}[t]\centering
\includegraphics[width=\linewidth]{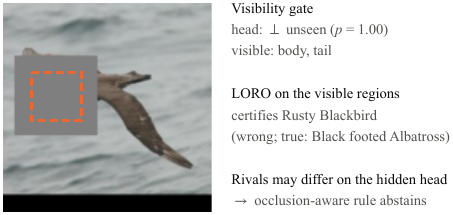}
\caption{CUB, unseen species, head grayed out (dashed box: ground-truth head keypoints). A separate visibility gate reports the head as unseen; a leave-one-region-out certificate on the remaining regions still certifies a wrong species, whereas the occlusion-aware rule (post hoc), which fills the hidden head with each rival's own attributes, abstains because a rival then wins. The image is the first test image, by index, that meets these conditions (96 of 1{,}460 qualify).}
\label{fig:s-cub-example}
\end{figure}
\paragraph{Occlusion-aware LORO.} Figure~\ref{fig:s-cub-example} shows an example. The gate reports 82\% of grayed-out heads and 81\% of heads covered by a background patch as unseen (1.1\% on clean images). The occlusion is a square over the ground-truth head keypoints, padded by 0.175 times the bird-box side on each side, applied before the keypoint model runs; the keypoint model still finds the head in 99.9\% of these images. For each rival, occlusion-aware LORO replaces the predicted attributes of every region that is not kept by the rival's class attributes, without rescaling, and recomputes the centered cosine of Sec.~\ref{sec:s-impl-cub} over the kept and the filled attributes; it certifies only if no rival then scores highest. With the head grayed out, the gate missed the head in 98\% of the images that occlusion-aware LORO still certifies, and none of the three region-based rules we tested matches APN-lite's small low-risk set. Agreement between regions reaches 19.4\% risk at the LORO coverage with the run-4 readers and 36.9\% with run 3's.

\paragraph{Development runs 1--3.} Table~\ref{tab:s-cubruns} lists the three runs that preceded run 4 on the same splits. Run 1's keypoint heatmaps did not separate visible from hidden regions; isolation only matched the joint reader (0.400 vs.\ 0.396) and reached 0.574 with ground-truth keypoints. Run 2 added a visibility head. Isolation then beat the joint reader by 17 points in every split, though not APN-lite in every split, and its risk fell with the number of regions whose attributes alone decode to the overall decision, from 58--68\% when none agreed to 4--10\% when four or more did. Because a grayed-out head was still marked visible in over 99\% of images, and APN-lite learned from class vectors rather than image-level labels, run 3 added an \textsc{absent} output to each reader and crossed the two supervisions. It flagged 87\% of grayed-out heads as unseen (0.2\% of clean heads), but the class-vector readers fell to 0.510. Run 4 moved $\bot$ into a separate gate and recovered 12 points with class-level attributes; run 3 differed in how it trained its readers (\textsc{absent} output, stricter positive set), and our runs do not separate the two.

\begin{table}[!tb]\centering\small
\caption{Matched controls on CUB (class-level attributes, clean images, retraining of run 4). All models use the same keypoints, gate, LORO rule, and test images. ISO: separate readers on $128^2$ crops of the $320^2$ bird image; SHARED: one network reads all five crops, with one head per region; ISO-224: separate readers on crops of the bird image reduced to $224^2$; POOL: features of one trunk on the $224^2$ image, pooled in the boxes ISO crops. Risk at ISO's LORO coverage (43.8\% on average). Shaded cells mark the best value per column at the shown precision.}
\label{tab:s-shared}
\setlength{\tabcolsep}{4pt}
\resizebox{\columnwidth}{!}{%
\begin{tabular}{lccc}
\toprule
Model & ZSL$\uparrow$ & Risk$\downarrow$ & Risk per split\\
\midrule
ISO & \best{0.643} & \best{18.6\%} & 13.8 / 20.8 / 21.2\\
SHARED & 0.629 & \best{18.6\%} & 16.1 / 18.4 / 21.3\\
ISO-224 & 0.631 & \best{18.6\%} & 14.6 / 20.1 / 21.2\\
POOL & 0.541 & 28.1\% & 26.5 / 26.4 / 31.4\\
\bottomrule
\end{tabular}}
\end{table}

\paragraph{Matched controls (post hoc).} The matched-keypoint comparison changes three things at once: whether a region is read from a crop or from pooled features, whether parameters are shared, and the image scale (Sec.~\ref{sec:cub}). Table~\ref{tab:s-shared} separates them on a retraining of run 4 that saved its keypoint model, gates, and readers, which the controls reuse. A single network that reads all five crops with one head per region matches the separate readers in risk and is 1.4 points lower in accuracy, and separate readers on crops of the bird image reduced to $224^2$, the resolution POOL sees, also reach the same risk. Pooling in the same boxes is 5.6--12.7 points riskier per split (9.5 on average). These controls do not separate the magnification of each region in its crop from the coarse feature map that POOL pools.

\begin{table}[!tb]\centering\small
\caption{CUB over all 200 species (rerun of run 4 that stored seen-class outputs; class-level attributes; seen and unseen test images pooled; mean over splits 0--2). $\gamma$: calibrated-stacking offset subtracted from seen-class scores, chosen per model to maximize $H$ on the test images (optimistic for both). $u$, $s$: per-class accuracy on unseen and seen species; all values are means of per-split values. Risk: LORO certificates of the isolated readers (ISO) and APN-lite's most confident images at ISO's coverage. Best $H$ maximizes over a grid of $\gamma$ and can differ in the third decimal from Table~\ref{tab:s-ausuc}, which maximizes over the whole seen--unseen curve.}
\label{tab:s-gzsl}
\setlength{\tabcolsep}{3pt}
\resizebox{\columnwidth}{!}{%
\begin{tabular}{lcccc}
\toprule
& $u$ / $s$ / $H$ ISO & $u$ / $s$ / $H$ APN-lite & Cov. & Risk ISO / APN-lite\\
\midrule
$\gamma=0$ & 0.251 / 0.737 / 0.375 & 0.313 / 0.708 / 0.433 & 46.7\% & 16.0\% / 15.1\%\\
calibrated & 0.547 / 0.593 / 0.569 & 0.467 / 0.593 / 0.523 & 43.7\% & 25.5\% / 24.8\%\\
\bottomrule
\end{tabular}}
\end{table}

\paragraph{All 200 species (post hoc).} Table~\ref{tab:s-gzsl} evaluates certificates as on FunnyBirds, among all classes and on pooled seen and unseen test images. Without calibration, both models favor seen species; the isolated readers then certify only 19.7\% of unseen images, at 74.2\% risk. With calibrated stacking, tuned on the test images for each model, the isolated readers reach the higher harmonic mean but certify at a risk similar to APN-lite's (per split 23.5, 20.3, and 32.6\% vs.\ 22.9, 24.4, and 27.2\%); on unseen images their certificates are wrong 23.7\% of the time at 35.2\% coverage. Occlusion-aware LORO over all 200 species, with the same offsets (post hoc), certifies 44.5\% of the pooled images at 15.2\% risk without calibration (APN-lite at that coverage: 14.2\%) and 41.6\% at 24.8\% with calibrated stacking (APN-lite: 24.1\%), and without calibration 73.2\% of its certificates on unseen images are wrong.

\paragraph{Bootstrap intervals (post hoc).} We resampled the 50 unseen species of each split 2{,}000 times (cluster bootstrap) and report 95\% intervals of the split-mean differences. In run 4 (splits 0--2), the isolated readers exceed APN-lite in ZSL accuracy by 7.9 points (3.0 to 12.6), APN-lite's LORO risk at the same coverage exceeds theirs by 6.4 points ($-0.1$ to 13.5), and the joint reader's exceeds theirs by 16.6 points (12.1 to 21.5). With matched keypoints, pooling trails the isolated readers by 9.4 accuracy points (7.1 to 12.0) and exceeds their risk by 11.5 points (7.8 to 15.1) on splits 0--2; on the held-out splits 3--5, the differences are 9.9 (7.9 to 12.1) and 9.7 points (6.7 to 13.0). In each resample, coverage and the competitors' top-ranked images at that coverage are recomputed, while each image's certificate is fixed. The intervals thus quantify evaluation uncertainty given the trained models; training variability is reported separately (run-to-run variability above).

\paragraph{Six splits and three training seeds (post hoc).} Because the intervals above hold the trained models fixed, and run 4 and its rerun differ, we added a comparison that includes training variability. We retrained APN-lite on all six splits of the matched-keypoint comparison (Table~\ref{tab:s-taskA}) and trained ISO and APN-lite twice more per split with other seeds, keeping the code and hyperparameters. With the ISO models of Table~\ref{tab:s-taskA} and the retrained APN-lite, LORO risk of APN-lite minus that of ISO is 6.4 points over six splits (95\% interval 1.9 to 11.3; splits 0--2: 0.7 to 14.2; splits 3--5: $-1.2$ to 13.0). A two-level bootstrap that resamples the species and, independently for each model and split, one of the three training seeds gives 6.3 points (1.1 to 11.5), whereas run 4's three-split interval above ($-0.1$ to 13.5) includes 0. The difference is positive in all 18 split--seed pairs (0.5 to 13.7 points; standard deviation over seeds within a split, 0.1 to 2.6 points), and ISO's ZSL accuracy exceeds APN-lite's in all 18 (61.6\% vs.\ 54.0\% on average). The retrained APN-lite differs from the run in Table~\ref{tab:s-taskA} by up to 4.8 risk points per split (mean 26.2\% vs.\ 27.0\%).

\begin{table}[!tb]\centering\small
\caption{SUB~\cite{bader2025sub}, base species seen in training (class-level attributes, retraining of run 4, mean over splits 0--2). Follow: share of images on which a model scores the substituted value above the base species' typical value of the same attribute type, per split in parentheses. Real: the same share on real CUB test images of the base species, which usually show the species-typical value (not verified per image). Add / remove: SUB's metrics, the share of images on which the substituted value scores at least 0.5 and on which the replaced value scores below 0.5. Head / other: follow rate by the region of the substituted attribute (grouped post hoc). Shaded cells mark the best value per column.}
\label{tab:s-sub}
\setlength{\tabcolsep}{3pt}
\resizebox{\columnwidth}{!}{%
\begin{tabular}{lccccc}
\toprule
Model & Follow$\uparrow$ & Real$\downarrow$ & Add / remove$\uparrow$ & Head$\uparrow$ & Other$\uparrow$\\
\midrule
Isolated readers & 55.9 (56.3, 57.2, 54.1) & \best{13.9} & 30.1 / 76.1 & 33.2 & \best{71.9}\\
Joint reader & 44.4 (44.2, 48.2, 40.8) & 15.8 & 20.0 / 73.3 & 32.9 & 52.5\\
APN-lite & \best{60.1} (62.3, 62.1, 55.9) & 31.6 & \best{35.6 / 76.5} & \best{57.3} & 62.0\\
\bottomrule
\end{tabular}}
\end{table}

\paragraph{Attribute substitutions (SUB, post hoc).} SUB~\cite{bader2025sub} contains 38{,}400 generated images of 33 CUB species, each with one attribute replaced (768 species--attribute pairs, 50 images each). We use the 29 substituted attributes that belong to one of our five regions (all except four primary-color values) and resize each $512^2$ image as a whole to the $320^2$ bird crop of run 4, since the bird is centered and fills most of the frame. For each image, we compare a model's score for the substituted value with its score for the value of the same attribute type that the base species shows most often in CUB (Table~\ref{tab:s-sub}). Region readers read crops at predicted keypoints; an undetected region counts as not following the substitution (fewer than 0.5\% of images), and the gate is not used. The models are a retraining of run 4 on splits 0--2 that also saved its weights (ZSL 0.643, LORO risk 18.6\% on unseen species), evaluated on the 24{,}900--26{,}750 images per split whose base species was seen in training. The isolated readers follow the substitution more often than the joint reader in every split, but less often than APN-lite. APN-lite, however, also ranks the other value higher on 31.6\% of real images of the species, which usually show the typical value. On the 41\% of images with a substituted head attribute, which one reader predicts together with 111 other attributes, the isolated readers follow it on 33.2\% of images, against 57.3\% for APN-lite, while on the other regions they lead both (71.9\% vs.\ 52.5\% and 62.0\%). With image-level labels, the isolated readers reach 62.8\%, and the joint reader reaches 52.3\%; on species unseen in training, the three models reach 56.7\%, 45.7\%, and 58.8\%.

\paragraph{Head sub-regions (post hoc).} We then split the head reader into five sub-region readers (bill; crown and forehead; eye; throat; nape), trained like the region readers on crops at the corresponding keypoints, with the same keypoint model, gate, and five-region LORO rule. This raises the head following rate on SUB from 33.2\% to 42.0\%, by 8.0--10.0 points in every split, while the rate on real images stays at 6.0\% (5.9\% before). It rises for the bill (68.1\% to 77.0\%), crown and forehead (42.8\% to 53.7\%), and throat (49.2\% to 62.7\%), but substituted eye colors are almost never followed (0.3\% and 1.7\%). On CUB, the sub-region readers raise ZSL accuracy from 0.643 to 0.670 and lower LORO risk at the same coverage from 18.6\% to 14.9\%, in every split. Most of the remaining head gap to APN-lite comes from eye color, 27\% of the head images: APN-lite ranks the substituted color higher on 77.5\% of SUB images, but also on 65.8\% of real images of the species, which usually show the typical color, against 0.0\% for the isolated and joint readers. Splitting the head also changes reader capacity and how training is divided, and this experiment does not separate these factors from shortcuts between attributes within one crop.

\section{Unseen traffic-sign compositions (MTSD)}\label{sec:s-mtsd}
\begin{table*}[t]\centering\small
\caption{Unseen traffic-sign compositions in MTSD (mean over 3 splits; 49 of 244 codes unseen per split). Cert.\ risk with coverage in parentheses, on signs of unseen codes. Shortcut: share of pictogram errors whose predicted pictogram occurs with the true frame among seen codes; in parentheses, the share expected if the wrong pictogram were drawn uniformly from the other pictograms (a uniform reference that does not account for pictogram frequency). The shortcut column counts argmax readings; the text also reports claims with a frequency reference. Pictogram acc.: per-sign accuracy of the pictogram claim argmax on unseen and seen codes. The two rows below the line come from a post hoc run with a circular content window (see text). Shaded cells mark the best value per column.}
\label{tab:s-mtsd}
\setlength{\tabcolsep}{4pt}
\resizebox{\textwidth}{!}{%
\begin{tabular}{lcccccc}
\toprule
Model & Cert.\ risk$\downarrow$ (cov.) & ZSL$\uparrow$ & GZSL$\uparrow$ & AUSUC$\uparrow$ & Shortcut$\downarrow$ (uniform ref.) & Pictogram acc.\ unseen / seen\\
\midrule
B2 joint CBM & 99.8\% (42.1\%) & 0.590 & 0.018 & 0.536 & 69.6\% (13.0\%) & 0.304 / 0.982\\
B2-ind (full-crop readers) & 92.2\% (38.5\%) & 0.680 & 0.109 & 0.630 & 67.4\% (12.9\%) & 0.373 / 0.980\\
\ocve{}-color (content reader sees color) & 81.4\% (28.4\%) & 0.674 & 0.167 & 0.630 & 53.9\% (12.2\%) & 0.355 / 0.980\\
\ocve{} (frame and content readers) & 67.2\% (32.5\%) & 0.747 & 0.317 & 0.701 & 35.5\% (12.5\%) & 0.384 / 0.968\\
\midrule
\ocve{}, circular window (ResNet-18) & 46.7\% (40.4\%) & 0.756 & 0.433 & 0.704 & \best{25.6\% (10.5\%)} & 0.441 / 0.952\\
\ocve{}, circular window (ViT-B/16) & \best{44.1\% (43.4\%)} & \best{0.780} & \best{0.489} & \best{0.728} & 27.3\% (10.2\%) & \best{0.455} / 0.957\\
\bottomrule
\end{tabular}}
\end{table*}

\paragraph{Data and ontology.} We use the fully annotated train and validation images of MTSD~\cite{ertler2020mtsd}, excluding panoramas. We keep signs that are not \texttt{other-sign}, not flagged as ambiguous, dummy, out-of-frame, exterior, or occluded, and whose shorter side is at least 32 px (35{,}917 training and 5{,}206 validation signs). Each sign is cropped as a square of side $1.2\max(w,h)$ around its box and resized to $128^2$. The concept table (Sec.~\ref{sec:protocol}) assigns every label a code over five groups: shape, border color, background color, pictogram (the symbol without its design-variant suffix), and text. The rule-based draft assumed Vienna-convention shapes and colors; checking each label against a montage of its validation crops changed 264 of the 400 labels and 728 of the 2{,}000 entries (214 shapes, 183 background colors, 178 border colors, 123 texts, and 30 symbols, 10 of which only merge design variants), for example yellow diamond warning signs and white rectangular regulatory plates used in the Americas and Australia. Each montage shows up to eight of the largest validation crops of a label, and validation signs also serve as test signs; the check used only these crops and the label names, no model output, and every model decodes with the resulting table. Groups whose color varies among a label's crops were marked as wildcards. The 12 labels without validation crops remain unchecked; they contribute 24 training signs, whose codes are seen in every split, and no test signs. Labels with identical codes are merged, and codes containing a wildcard or fewer than 30 signs are removed, leaving 244 codes from 286 labels, with 10, 6, 7, 141, and 44 values per group.

\paragraph{Splits.} Each of three splits (seeds 0--2) holds out 49 codes (20\%) as unseen, drawn at random among codes whose every value also occurs in some other code, and accepted only if every value of every unseen code remains present among the seen codes. Training uses the training-set signs of seen codes minus 10\% held out to calibrate $\tau$; the unseen test set contains all signs of unseen codes (7{,}511--10{,}152 per split), and the seen test set contains the validation-set signs of seen codes (3{,}227--3{,}549).

\paragraph{Models.} All models use an ImageNet ResNet-18 at $128^2$, 8 epochs, AdamW ($3\cdot10^{-4}$, one-cycle schedule), batch size 128, and random translation by up to 8 px; flips are not used, since left and right carry meaning. The joint CBM (B2) predicts the five groups with five heads on the full crop, and B2-ind trains one network per group on the full crop. \ocve{} uses two isolated readers: a frame reader sees the crop with its central region (pixels 32--96) filled with the ImageNet mean color and predicts shape, border, and background; a content reader sees only the central region, upsampled to $128^2$ and converted to grayscale, with random polarity inversion during training, and predicts pictogram and text. The ablation \ocve{}-color keeps the frame reader and gives the content reader the central region in color, without polarity inversion. Claims, $\tau$ (1\% quantile of the maximum probability of correct claims on the calibration set), and set decoding follow FunnyBirds; a certificate requires that the claims match exactly one of all 244 codes.

\paragraph{Results.} Table~\ref{tab:s-mtsd} gives the results. The joint CBM certifies 42\% of unseen-code signs and is wrong on 99.8\% of them (99.7--99.9\% per split), while on seen codes every model certifies 94--97\% of signs at about 1\% risk. Among the joint CBM's pictogram errors on unseen codes, 69.6\% predict a pictogram that occurs with the sign's true frame among seen codes, against 13.0\% expected if the wrong pictogram were drawn from the other pictograms. \ocve{} halves the shortcut rate (35.5\%) and the number of wrong certificates (5{,}484 vs.\ 10{,}523 over three splits) and raises ZSL from 0.590 to 0.747 and AUSUC from 0.536 to 0.701, yet 67.2\% of its certificates are wrong (56.3--74.2\% per split). Replacing the grayscale, polarity-invariant content reader with its color counterpart raises its risk to 81.4\% and its pictogram shortcut rate to 53.9\%, in every split; the two readers differ in both color and augmentation. Because the uniform reference ignores how often pictograms occur, we also counted claims only ($\bot$ excluded, splits pooled): 84.4\% of the joint CBM's wrong pictograms and 31.6\% of \ocve{}'s occur with the true frame, against 17.5\% and 13.3\% expected if the wrong pictogram followed pictogram frequencies among validation signs of seen codes.

\paragraph{Where the remaining errors come from (post hoc).} Of \ocve{}'s wrong certificates, 98\% name a seen code. The certified code differs from the true one only in the pictogram in 47\% of them, in background and pictogram in 13\%, and only in shape or only in text in 8\% each; for the joint CBM, the pictogram alone accounts for 54\%. On unseen codes, nearly all certificates of every model have $d(x)=1$, and the few with $d(x)=2$ are all wrong (29 over three splits for \ocve{}, 24 for the joint CBM, and 37 for B2-ind). Of the 244 codes, 218 have another code at distance 1, including all unseen codes in two splits and all but one in the third, so a correct certificate of an unseen sign almost always has $d(x)\le1$. A certificate with $d(x)=2$ must name one of the 26 codes without a distance-1 neighbor, which are almost never the true code of an unseen sign, and arises only when several claims are wrong together. In \ocve{}'s 29 cases, the misread groups include pictogram and text, both predicted by the content reader, in 16 cases, shape and pictogram in 7, and two groups of the frame reader in 6 (10 of the 29 have a third misread group); $d(x)$ counts concept groups rather than independent readers. On seen codes, which include the codes without a distance-1 neighbor, \ocve{}'s certificates are wrong in 2 of 1{,}401 cases at $d(x)\ge2$ and in 105 of 8{,}265 at $d(x)=1$ (joint CBM: 5 of 1{,}521 and 97 of 8{,}481). Pictogram accuracy on unseen codes stays low for every model (0.30--0.38, compared with 0.97--0.98 on seen codes).

\paragraph{Reader-level distance and set-valued certificates (post hoc).} Because $d(x)$ counts groups, we also counted, for each rival, the readers with at least one claim that excludes it (frame and content for \ocve{}). On unseen codes, all 14 of \ocve{}'s certificates at reader distance 2 are wrong (group distance 2: 29 of 29), whereas on seen codes 2 of 1{,}048 are wrong (reader distance 1: 105 of 8{,}618). Accepting compatible sets of up to 2, 3, or 5 codes barely lowers the risk on unseen codes (\ocve{}: 64.6, 64.8, and 64.4\% at 34.7--36.6\% coverage; joint CBM: 98.4, 96.5, and 95.2\%), because a misread pictogram usually excludes the true code from the set. The 5\% of \ocve{}'s certificates of unseen codes with the highest posterior are wrong 47\% of the time.

The remaining shortcut rate differs by sign shape. The central region given to the content reader contains the slanted edges of triangular and diamond-shaped signs but lies inside square ones. Among \ocve{}'s pictogram claims on unseen codes (pooled over splits, $\bot$ excluded), 93.5\% of the errors on upward triangles and 61.8\% on diamonds predict a pictogram that occurs with the true frame (uniform reference 15.6\% and 30.2\%), against 16.1\% on circles and 0.7\% on squares (7.5\% and 5.5\%). Without triangles and diamonds, pooled risk changes only from 65.8\% to 64.0\%, and certificates on circles alone are wrong in 80.8\% of cases.

\paragraph{A content window without frame edges (post hoc).} After observing this pattern, we reran \ocve{} once per configuration with a content window of the same shape for every sign: a disk inside the plate, clear of its border, with the rest of the crop grayed before upsampling to $128^2$. For triangles, the disk is centered at the incenter and has a radius of 0.62 times the inradius; for diamonds, its radius is 0.8 times the inradius; for other shapes, it is the disk of radius 32\,px inscribed in the former window. Training places the window with the true shape, and testing places it with the shape predicted by the frame reader. The frame reader's input, the codes, and the splits are unchanged. With ResNet-18 readers, certified risk on unseen codes falls from 67.2\% to 46.7\% (57.3, 40.8, and 41.9\% per split) at 40.4\% coverage, and with ViT-B/16 readers (inputs resized to $224^2$, AdamW at $10^{-4}$, batch size 64), it falls to 44.1\% (56.2, 40.4, and 35.6\%) at 43.4\% coverage (Table~\ref{tab:s-mtsd}). The window therefore accounts for 20.5 of the 23.1 points. With ViT-B/16 readers, correct certificates rise from 2{,}850 to 6{,}393 over three splits, and the pictogram shortcut rate among claims falls to 27.1\% against a frequency reference of 12.9\% (ResNet-18: 24.3\% vs.\ 12.6\%). On seen codes, risk stays at 1.05--1.25\% (92.7--93.6\% coverage). On unseen codes, 44\% of the ViT-B/16 certificates are wrong, the 5\% with the highest posterior are wrong in 14.5--50.0\% of cases per split, and all 36 certificates with $d(x)=2$ are wrong.

\paragraph{Which pictograms fail (post hoc).} We grouped the pictogram claims on unseen-code signs (three splits pooled) by three properties of the sign. The clearest factor is how often the pictogram occurs in training: for pictograms with more than 259 training signs (top tertile), 26\% of \ocve{}'s claims are wrong, against 67--77\% in the lower two tertiles, and the joint CBM and \ocve{} with the circular window (ViT-B/16) show the same drop (21 and 27\% vs.\ 58--76\%). Wrong-claim rates hardly change across tertiles of sign size (47--55\% for all three models). Whether the true pictogram occurs with the sign's shape among seen codes changes the wrong-claim rate by 26 points for \ocve{} and 70 points for the joint CBM, whereas the gap shrinks to 8 points with the circular window, consistent with frame edges in the square window acting as cues. The most frequent remaining confusions are fine distinctions on rare pictograms, such as left and right chevrons or winding roads (although training uses no flips), parking and no-parking plates, and an ambiguity of the ontology between square and upright rectangular plates. With the circular window (ViT-B/16), the three codes with the most wrong certificates contribute 39--51\% of them per split, and 16--22 of the 48--49 unseen codes with certificates are wrong at least 90\% of the time.

\paragraph{Design variants as separate pictograms (post hoc).} The pictogram values above merge the design variants of a symbol. To test whether this merging contributes to the pictogram errors, we kept the variants separate, which gives 265 codes with 217 pictograms. We redrew the splits with the same rules (25 unseen codes per split; 3{,}188--4{,}533 unseen-code signs) and trained \ocve{} with the circular window and a joint CBM, both with ViT-B/16 readers. The joint CBM certifies 29--50\% of unseen-code signs per split and is wrong on all 4{,}650 of these certificates, each of which names a seen code; 76.3\% of its wrong pictogram claims occur with the true frame, against 8.2\% expected from pictogram frequencies. \ocve{} is wrong on 45.0\% of its certificates (50.6, 26.5, and 57.9\% per split) at 37.0\% coverage, with a pictogram shortcut rate of 13.5\% (9.5\% expected), and it reaches higher ZSL accuracy (0.772 vs.\ 0.664) and AUSUC (0.730 vs.\ 0.628). Its risk per code is higher (68--80\%), because a few large codes contribute most correct certificates; in split~1, one code accounts for 1{,}056 of 1{,}403. Pictogram accuracy on unseen codes is 0.610 for \ocve{} and 0.663 for the joint CBM. Of \ocve{}'s wrong certificates, 58\% differ from the true code only in the text, mostly in the number of a speed limit, and 27\% differ only in the pictogram. On seen codes, risk is 1.5\% for \ocve{} and 0.9\% for the joint CBM at 92--98\% coverage.

\section{Seen classes: corruption study}\label{sec:s-seen}

\begin{table}[t]\centering\small
\caption{Seen FunnyBirds classes under nine image corruptions (means over the conditions). Risk: certified risk of \ocve{}'s certificates with $d(x)\ge2$, and B1's risk at the same coverage when ranked by confidence. AUROC: how well each model's score ranks correct above wrong predictions. Clean acc.: accuracy on uncorrupted images (\ocve{}: soft decoder).}
\label{tab:s-seen-corr}
\setlength{\tabcolsep}{3pt}
\resizebox{\columnwidth}{!}{%
\begin{tabular}{lccc}
\toprule
& oracle masks & learned localizer & + fixes, shared aug.\\
\midrule
Risk (\%)$\downarrow$ \ocve{} / B1 & 0.0 / 21.2 & 19.7 / 2.3 & 0.0 / 0.1\\
AUROC$\uparrow$ \ocve{} / B1 & 0.963 / 0.844 & 0.813 / 0.844 & 0.952 / 0.944\\
Clean acc.$\uparrow$ \ocve{} / B1 & 0.785 / 0.709 & 0.723 / 0.709 & 0.671 / 0.735\\
\bottomrule
\end{tabular}}
\end{table}

\paragraph{Corruption study on seen classes.} Sec.~\ref{sec:seen} summarizes Table~\ref{tab:s-seen-corr}. All 50 FunnyBirds classes are seen; models train on 12{,}000 images, set thresholds on 1{,}000 validation images, and are tested on 2{,}500 images (the 500 official test images and 2{,}000 held-out training-split images). Each corruption is applied to the full image before any model sees it: additive Gaussian noise with standard deviation 0.04, 0.08, or 0.16 on pixel values in $[0,1]$, Gaussian blur with radius 1, 2, or 3\,px, or contrast reduced toward the image mean by a factor of 0.6, 0.4, or 0.25. In the first setting, the readers are trained and tested on crops from the ground-truth part maps, and B1 is a ResNet-18 on the full image, each trained for 3 epochs. The second setting keeps these models and takes the test crops from a part segmenter trained for 4 epochs. In the third, the segmenter, the readers (now with an \textsc{absent} class and trained on crops from predicted masks), and B1 are retrained for 5 epochs with a shared augmentation: with probability 0.5, an image receives one corruption of random type at a continuous level (standard deviation 0.02--0.16, radius 0.5--3\,px, or factor 0.25--0.9). Risk is computed per condition and averaged over the nine conditions; B1 ranks its predictions by the maximum softmax probability. AUROC is the area under the ROC curve for separating correct from wrong predictions, with the maximum softmax probability for B1 and, for \ocve{}, $d(x)$ followed by the soft decoder's confidence. Over the 50 classes, the Spearman correlation between a class's code distance to its nearest other class and its rate of wrong certificates under corruption is $-0.13$, $-0.17$, and $+0.14$ in the three settings.

\end{document}